\documentclass{article} 
\usepackage{iclr2026_conference,times}
\usepackage{hyperref}
\usepackage{url}
\usepackage{mathtools}
\usepackage{cleveref}
\usepackage{multirow}
\usepackage{booktabs}
\usepackage{stmaryrd}
\usepackage{enumitem}
\usepackage{graphicx}   
\usepackage{subcaption}
\usepackage{url}
\usepackage{array}
\usepackage{booktabs}
\usepackage{colortbl}
\usepackage[table]{xcolor}
\usepackage{listings}
\usepackage{wrapfig}
\usepackage{algorithm}
\usepackage{xspace}
\usepackage[inkscapelatex=false]{svg}
\usepackage{makecell}
\usepackage{amssymb}
\usepackage{tcolorbox}
\usepackage{stmaryrd}

\newcolumntype{x}[1]{>{\centering\arraybackslash}p{#1pt}}
\newcolumntype{L}[1]{>{\raggedright\arraybackslash}p{#1pt}}
\newcommand{\tablestyle}[2]{%
  \setlength{\tabcolsep}{#1}%
  \renewcommand{\arraystretch}{#2}%
}
\definecolor{baselinecolor}{gray}{0.85}

\usepackage{ifthen} 
\newcommand{\expect}[2][]{
\ifthenelse{\equal{#1}{}}{
\mathbb{E}\left[#2\right]
}{
\underset{#1}{\mathbb{E}}\left[#2\right]
}}
\renewcommand{\arraystretch}{1.5}

\usepackage{amsmath,amsfonts,bm,amsthm}

\def\1{\bm{1}}

\def\vf{{\bm{f}}}

\def\vu{{\bm{u}}}
\def\vv{{\bm{v}}}

\def\vx{{\bm{x}}}

\def\vz{{\bm{z}}}
\def\veps{{\bm{\epsilon}}}

\DeclareMathAlphabet{\mathsfit}{\encodingdefault}{\sfdefault}{m}{sl}
\SetMathAlphabet{\mathsfit}{bold}{\encodingdefault}{\sfdefault}{bx}{n}

\def\gN{{\mathcal{N}}}

\newtheorem{thm}{Theorem}

\newtheorem{defi}{Definition}

\crefname{thm}{theorem}{theorems}
\Crefname{thm}{Theorem}{Theorems}
\crefname{obs}{observation}{observation}
\Crefname{obs}{Observation}{Observations}
\crefname{prop}{proposition}{propositions}
\Crefname{prop}{Proposition}{Propositions}

\newcommand{\FID}{FID\xspace}
\newcommand{\FDD}{FDD\xspace}

\newcommand{\FCD}{FCD\xspace}

\newcommand{\ditxl}{DiT-XL/2\xspace}
\newcommand{\ditxlp}{DiT-XL/2+\xspace}
\newcommand{\ie}{\emph{i.e.}}

\newcommand{\stepratio}{\alpha}

\newcommand{\omegaMFd}{\frac{t - r}{\Delta t}}

\newcommand{\LFM}{\ensuremath{\mathcal{L}_{\mathtt{FM}}}\xspace}

\newcommand{\LCFc}{\ensuremath{\mathcal{L}_{\mathtt{TC_c}}}\xspace}
\newcommand{\LuFM}{\ensuremath{\mathcal{L}_{\mathtt{TFM}}}\xspace}
\newcommand{\LbFM}{\ensuremath{\mathcal{L}_{\mathtt{FM'}}}\xspace}
\newcommand{\LCTc}{\ensuremath{\mathcal{L}_{\mathtt{CT_c}}}\xspace}
\newcommand{\LCTd}{\ensuremath{\mathcal{L}_{\mathtt{CT_d}}}\xspace}
\newcommand{\LSC}{\ensuremath{\mathcal{L}_{\mathtt{SC}}}\xspace}
\newcommand{\LMFd}{\ensuremath{\mathcal{L}_\alpha}}

\newcommand{\LMFc}{\ensuremath{\mathcal{L}_{\mathtt{MF}}}\xspace}
\newcommand{\omegacfg}{w}
\newcommand{\kappacfg}{\kappa}

\newcommand{\param}{\bm{\theta}}
\newcommand{\paramsg}{\param^{-}}
\newcommand{\vutheta}{\vu_{\param}}
\newcommand{\vuthetasg}{\vu_{\paramsg}}
\newcommand{\vvoffset}{\tilde{\vv}_{s,t}}
\newcommand{\rationame}{consistency step ratio\xspace}
\newcommand{\Rationame}{Consistency step ratio\xspace}
\newcommand{\modelname}{$\stepratio$\text{-Flow}\xspace}
\newcommand{\modelnamexlp}{$\stepratio$\text{-Flow}-XL/2+\xspace}

\newcommand{\sname}{intermediate timestep\xspace}
\newcommand{\INFull}{ImageNet-1K $256^2$\xspace}

\newcommand{\inlinesection}[1]{\noindent \paragraph{#1.}}
\newcommand{\bFMname}{flow matching\xspace}
\newcommand{\uFMname}{trajectory flow matching\xspace}
\newcommand{\BFMname}{flow matching\xspace}
\newcommand{\UFMname}{Trajectory flow matching\xspace}
\newcommand{\CFname}{Trajectory consistency\xspace}
\newcommand{\cFname}{trajectory consistency\xspace}
\newcommand{\ours}{\textcolor{azure}{(ours)}}
\newcommand{\SigmoidSch}[2]{\ensuremath{\mathtt{Sigmoid}_{#1\mathrm{K} \rightarrow #2\mathrm{K}}}}
\newcommand{\Frechet}{Fréchet\xspace}
\newcommand{\emptytable}{--}

\newcommand{\vzzerohat}{\hat{\bm{z}}_0}

\newcommand{\vzt}{\bm{z}_t}
\newcommand{\vzs}{\bm{z}_s}
\newcommand{\vvt}{\bm{v}_t}
\newcommand{\vzzero}{\bm{z}_0}

\newcommand{\expecttrzt}[1]{\ensuremath{\expect[t,r,\vz_t]{#1}}}

\usepackage{textcomp}  
\usepackage{scalerel}  

\definecolor{mediumtealblue}{rgb}{0.0, 0.33, 0.71}
\definecolor{darkpastelgreen}{rgb}{0.01, 0.75, 0.24}
\definecolor{azure}{rgb}{0.0, 0.5, 1.0}

\definecolor{crimsonred}{rgb}{0.86, 0.08, 0.24}
\definecolor{firebrick}{rgb}{0.7, 0.13, 0.13}
\definecolor{carmine}{rgb}{0.59, 0.0, 0.09}
\definecolor{rosewood}{rgb}{0.4, 0.0, 0.04}
\definecolor{deepcherry}{rgb}{0.6, 0.13, 0.13}

\definecolor{codeblue}{rgb}{0.25,0.5,0.5}
\definecolor{codekw}{rgb}{0.85, 0.18, 0.50}
\definecolor{codesign}{RGB}{0, 0, 255}
\definecolor{codefunc}{rgb}{0.85, 0.18, 0.50}
\definecolor{lightyellow}{rgb}{1.0, 1.0, 0.8}

\lstdefinelanguage{PythonFuncColor}{
  language=Python,
  keywordstyle=\color{black},
  commentstyle=\color{codeblue},  
  stringstyle=\color{orange},
  showstringspaces=false,
  basicstyle=\ttfamily\small,
  literate=
    {*}{{\color{codesign}* }}{1}
    {-}{{\color{codesign}- }}{1}
    {+}{{\color{codesign}+ }}{1}
    {dataloader}{{\color{codefunc}dataloader}}{1}
    {sample_t_r}{{\color{codefunc}sample\_t\_r}}{1}
    {sample_alpha}{{\color{codefunc}sample\_alpha}}{1}
    {randn}{{\color{codefunc}randn}}{1}
    {randn_like}{{\color{codefunc}randn\_like}}{1}
    {jvp}{{\color{codefunc}jvp}}{1}
    {exp}{{\color{codefunc}exp}}{1}
    {if}{{\color{codefunc}if }}{1}
    {else}{{\color{codefunc}else }}{1}
    {elif}{{\color{codefunc}elif }}{1}
    {stopgrad}{{\color{codefunc}stopgrad}}{1}
    {sigmoid}{{\color{codefunc}sigmoid}}{1}
    {metric}{{\color{codefunc}metric}}{1}
    {range}{{\color{codefunc}range}}{1}
    {consistency_sampling}{{\color{codefunc}consistency\_sampling}}{1}
    {ODE_sampling}{{\color{codefunc}ODE\_sampling}}{1}
}

\newcommand{\todo}[2][]{
\ifthenelse{\equal{#1}{}}{
\textcolor{red}{TODO: #2}
}{
\textcolor{red}{TODO \textcolor{blue}{[#1]}: #2}
}}

\title{AlphaFlow: Understanding and Improving MeanFlow Models}

\author{Huijie Zhang $^{1, 2 }$ \thanks{Work done during an internship at Snap Inc.} \quad Aliaksandr Siarohin$^{1}$ \quad Willi Menapace$^{1}$ \\
\textbf{Michael Vasilkovsky$^{1}$ \quad Sergey Tulyakov$^{1}$ \quad Qing Qu$^{2}$ \quad Ivan Skorokhodov$^{1}$} \\
$^1$Snap Inc. \  $^2$Department of EECS, University of Michigan\\
}

\iclrfinalcopy 
\begin{document}

\maketitle

\begin{figure}[htbp]
\centering
\vspace{-0.2in}
\includegraphics[width=\linewidth]{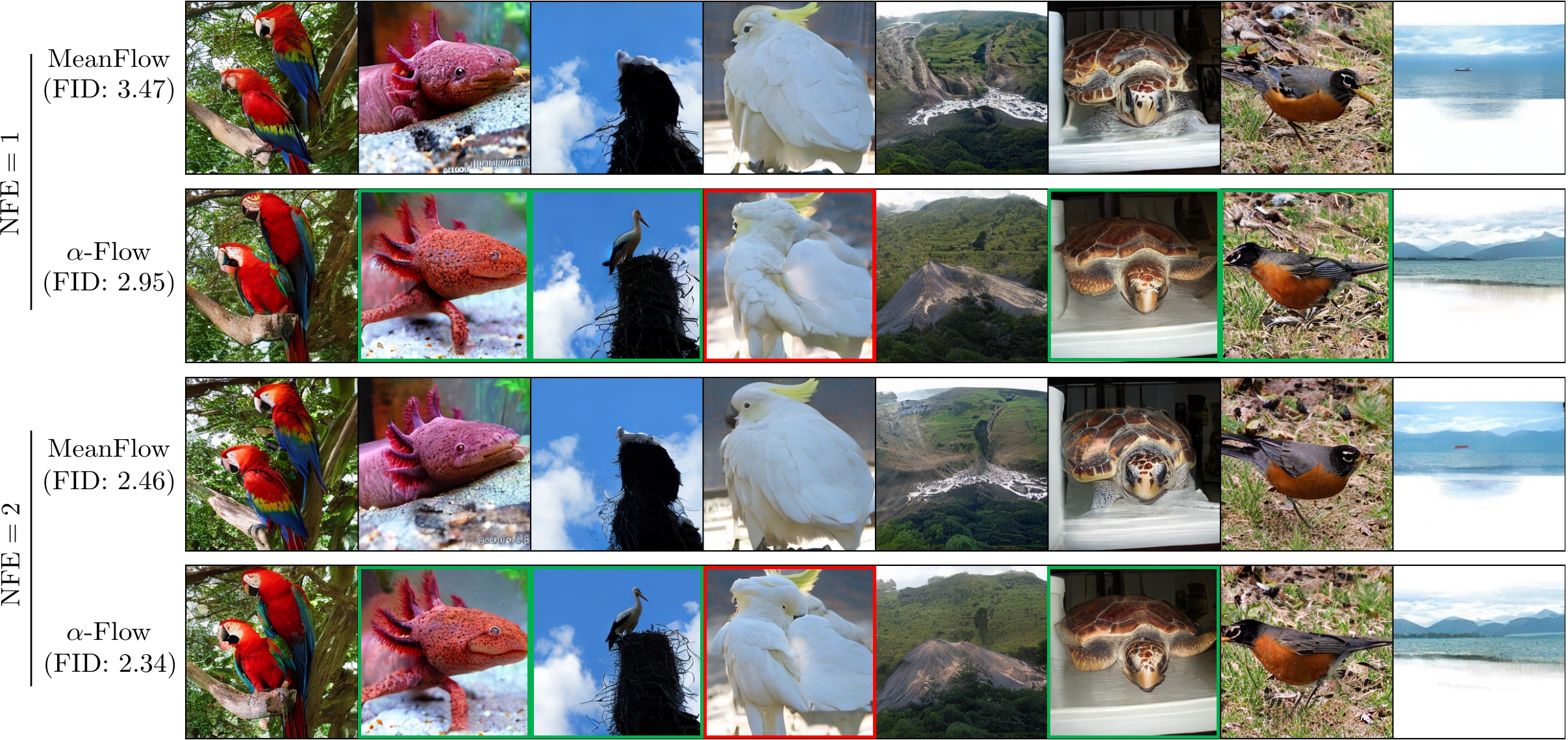}
\caption{\emph{Uncurated} samples (seeds $1$-$8$) from the \ditxl model for MeanFlow~\cite{MeanFlow} and \modelname (our proposed method) produced with 1 (upper) and 2 (lower) sampling steps for \INFull.}
\vspace{-0.1in}
\label{fig:teaser}
\end{figure}

\begin{abstract}

MeanFlow has recently emerged as a powerful framework for few-step generative modeling trained from scratch, but its success is not yet fully understood.
In this work, we show that the MeanFlow objective naturally decomposes into two parts: trajectory flow matching and trajectory consistency.
Through gradient analysis, we find that these terms are strongly negatively correlated, causing optimization conflict and slow convergence.
Motivated by these insights, we introduce \modelname, a broad family of objectives that unifies trajectory flow matching, Shortcut Model, and MeanFlow under one formulation.
By adopting a curriculum strategy that smoothly anneals from trajectory flow matching to MeanFlow, \modelname disentangles the conflicting objectives, and achieves better convergence.
When trained from scratch on class-conditional ImageNet-1K 256×256 with vanilla DiT backbones, \modelname consistently outperforms MeanFlow across scales and settings.
Our largest \modelname-XL/2+ model achieves new state-of-the-art results using vanilla DiT backbones, with FID scores of 2.58 (1-NFE) and 2.15 (2-NFE).
The source code and pre-trained checkpoints are available on \url{https://github.com/snap-research/alphaflow}.


\end{abstract}

\section{Introduction}
\label{sec:intro}

Diffusion models~\citep{NonEqThermodynamics} have emerged as the leading paradigm for generative modeling of visual data~\citep{ADM,LDM,Sora}.
However, their widespread use is limited by slow inference, as generating high-fidelity samples typically requires a large number of denoising steps.
This computational bottleneck has spurred extensive research into designing efficient diffusion-based generators that are able to operate in very few steps while preserving high generation quality~\citep{ProgressiveDistillation, ADD, CMs, iCT, sCM, EasyCM, ShortcutModels, MeanFlow}.

Early attempts reduce the inference time of diffusion models through distilling a pre-trained multi-step model into a few-step one \citep{ProgressiveDistillation,ADD}.
The subsequent development of consistency models \citep{CMs,iCT,sCM} enabled training from scratch for few-step generative models. 
However, a significant performance gap still remains between existing few-step and multi-step diffusion models.
The recently introduced MeanFlow framework \citep{MeanFlow} enables more stable training and better classifier-free guidance~\citep{CFG} integration, significantly bridging the gap between few-step and multi-step from-scratch trained diffusion models. 
Despite its practical success, there still lacks a clear understanding of why MeanFlow performs better, which hinders further improvements and the design of stronger few-step models.

In this work, we provide a deeper understanding of why MeanFlow works, revealing that its training objective can be decomposed into two components: trajectory flow matching and trajectory consistency.
Our gradient analysis shows that these two components are strongly negatively correlated during training, leading to instability and slow convergence in joint optimization.
We further demonstrate that the previous heuristic adoption of border-case flow matching supervision is crucial: it actually acts as a surrogate loss for trajectory flow matching and mitigates gradient conflict.
However, over 75\% of MeanFlow’s computation is spent on this border-case supervision, which is not its primary focus. 
This raises an open question: \emph{can we design more efficient techniques to optimize MeanFlow objective, without such computational overhead?}

Motivated by these observations, we introduce \modelname, a new broad family of objectives for few-step flow models.
This framework unifies trajectory flow matching, Shortcut Models \cite{ShortcutModels}, and MeanFlow under a single unified formulation.
By employing a curriculum learning strategy that smoothly transitions from trajectory flow matching to MeanFlow, \modelname better disentangles the optimization of trajectory flow matching and trajectory consistency, reduces reliance on border-case flow matching supervision, and achieves better convergence.

By training vanilla DiT-\citep{DiT} models from scratch with $\alpha$-Flow on class-conditional \INFull, we obtain consistently stronger performance across both small- and large-scale settings compared with MeanFlow, for both one-step and few-step generation. Our largest \ditxlp model establishes new state-of-the-art results among all from-scratch trained models with the vanilla DiT backbone and training pipeline, achieving FID scores of $2.58$ (1-NFE) and $2.15$ (2-NFE).

\section{Preliminaries}
\label{sec:prelim}


\inlinesection{Diffusion models and flow matching} Diffusion model \citep{DDPM, NCSN, LDM} define a forward process that progressively adds noise to a data sample $\vx \sim p_{\mathtt{data}}(\vx)$ over a continuous timestep $t\in[0, 1]$. 
Specifically, given training data, the forward process perturbs $\vx$ into a noisy version $\vzt = \beta_t \bm x + \sigma_t \bm \epsilon$ where $\veps \sim \gN(\bm 0, \bm I)$, $\beta_t$ and $\sigma_t$ are pre-defined scheduler parameters that depend on $t$, such that $\bm z_0 = \bm x$ and $\bm z_1 = \bm \epsilon$.
Flow matching \citep{FlowStraightAndFast, flowmatching} is a deterministic alternative that defines the forward process as a straight-line path between the noise distribution and the data distribution, setting $\beta_t = 1 -t$ and $\sigma_t = t$.
A neural network $\vv_{\param}(\vzt, t)$ is trained to model the ground-truth vector field $\text{d} \vzt / \text{d} t$ along this trajectory $\vzt$ by minimizing the objective: 
\begin{equation}\label{eq:flow_matching_loss}
\LFM\left(\param \right) =  \mathbb{E}_{t, \vx, \vzt} [||\vv_{\param}(\vzt, t) - \vvt ||^2]
\end{equation}
where $\vvt \triangleq \vv(\vzt, t|\vx) = \text{d} \vzt / \text{d} t \big|_{\vx} = \bm \epsilon - \bm x$.
To generate a new sample, the probability flow ODE (PF-ODE) $\text{d} \vz / \text{d} t = \vv_{\param}(\vzt, t)$ is solved from $t = 1$ to $t = 0$, starting with an initial value $\vz_1 \sim \gN(\bm 0, \bm I)$. 

One primary challenge of diffusion models is the slow sampling speed.
To address this, several methods have been proposed to enable high-quality generation with significantly fewer steps.

\inlinesection{Consistency model (CM)} \citep{CMs} enables one-step generation by training a neural network $\vf_{\param}(\vzt, t)$ to directly map the noisy input $\vzt$ to clean samples $\bm x$.
The core idea is to enforce a consistency property at any two nearby timesteps $t$ and $s$, by minimizing the difference between the model's output.
Depending on the $\Delta t \coloneqq t - s$, the training objective can be categorized into:
\begin{itemize}[leftmargin=*]
\item \textit{Discrete-time Consistency Training (CT)} \citep{EasyCM, CMs, iCT} minimizes the following discrete time CT loss $\mathcal{L}_{\mathtt{CT_d}}$:
\begin{equation}\label{eq:discrete_consistency_loss}
\LCTd(\param) =\mathbb{E}_{t, s, \vzt}\left[\left\Vert \vf_{\param}(\vzt, t) - \vf_{\paramsg}\left(\vz_s, s\right)\right\Vert^2_2\right],
\end{equation}
where $0 \leq s < t \leq 1$, $\vz_s = \vzt - \Delta t \cdot \vv$ and $\vf_{\paramsg} \coloneqq \mathtt{stopgrad}\left(\vf_{\param}\right)$.
While smaller values of $\Delta t$ reduce the discretization error and improve performance, they might also lead to training instability \citep{CMs, EasyCM}.
This necessitates a carefully designed scheduler for $\Delta t$ to ensure good performance and stability during training.

\item \textit{Continuous-time CT} \citep{sCM, CMs} eliminates the discretization error by the continuous time CT loss $\mathcal{L}_{\mathtt{CT_c}}$:
\begin{equation}\label{eq:continuous_consistency_loss}
\LCTc(\param) =2 \mathbb{E}_{t, \vzt}\left[\vf^{\top}_{\param}(\vzt, t) \frac{\text{d} \vf_{\paramsg}\left(\vzt,  t\right)}{\text{d} t}\right],
\end{equation}
\cite{CMs} theoretically show that $\nabla_{\param} \mathcal{L}_{\mathtt{CT_c}}(\param) = \lim_{\Delta t \rightarrow 0} \nabla_{\param} \mathcal{L}_{\mathtt{CT_d}}(\param)/\Delta t$.
However, estimating $\frac{\text{d} \vf_{\paramsg}\left(\vzt,  t\right)}{\text{d} t}$ relies on the Jacobian-vector product (JVP) operation, which causes potential issues of scalability and efficiency in modern deep learning frameworks \citep{TransitionModels,FACM}.
\end{itemize}
\inlinesection{Consistency trajectory model (CTM)} \citep{CTM, IMM, ShortcutModels, MeanFlow} generalize Consistency Models (CMs) by training a neural network $\bm u_{\param}(\vzt, r, t)$ to enforce consistency across a trajectory from $t$ to $r$ with $0 \leq r \leq t \leq 1$.
This allows jumping from any $t \in (0,1]$ to any $r < t$ during inference, enabling multi-step generation.
To train CTM from scratch:
\begin{itemize}[leftmargin=*]
\item \emph{Shortcut model} \citep{ShortcutModels} enforces consistency by ensuring that a single "shortcut" step from $t$ to $r$ is consistent with two consecutive shortcut steps of half the size.
The training objective is:

\begin{equation}
\begin{aligned}
\LSC(\param) &=\expecttrzt{\left\Vert\vu_{\param}(\vzt, r, t) - \vu_{\paramsg}(\vzt, s, t)/2 - \vu_{\paramsg}\left(\vz_{s}, r, s\right)/2\right\Vert^2_2}, \\
\end{aligned}
\end{equation}
where $\vz_{s} = \vzt - \left(t - s\right) \cdot \vu_{\paramsg}(\vzt, s, t)$ and $s = (t + r)/2$.
\item \textit{MeanFlow} \citep{MeanFlow} trains the model $\bm u_{\param}(\vzt, r, t)$ to estimate the mean velocity $\frac{1}{t -r}\int_{r}^{t}\vv(\vz_\tau, \tau)\text{d} \tau$, with training objective given by:

\begin{equation}\label{eq:mean_flow_loss}
\LMFc(\param) = \expecttrzt{\left\Vert\vu_{\param}(\vzt, r, t) - \vvt + (t - r) \frac{\text{d} \vu_{\paramsg}(\vzt, r, t)}{\text{d} t} \right\Vert^2_2}.
\end{equation}

\end{itemize}

In practice, MeanFlow significantly outperforms other one/few-step diffusion and flow models. Yet, there has been little analysis explaining why it works so effectively.
To shed light on this, we analyze MeanFlow training in the next section.

\section{Analyzing MeanFlow Training}
\label{sec:analysis}
\begin{figure}
\centering
\begin{subfigure}{0.32\textwidth}
\includegraphics[width=\linewidth]{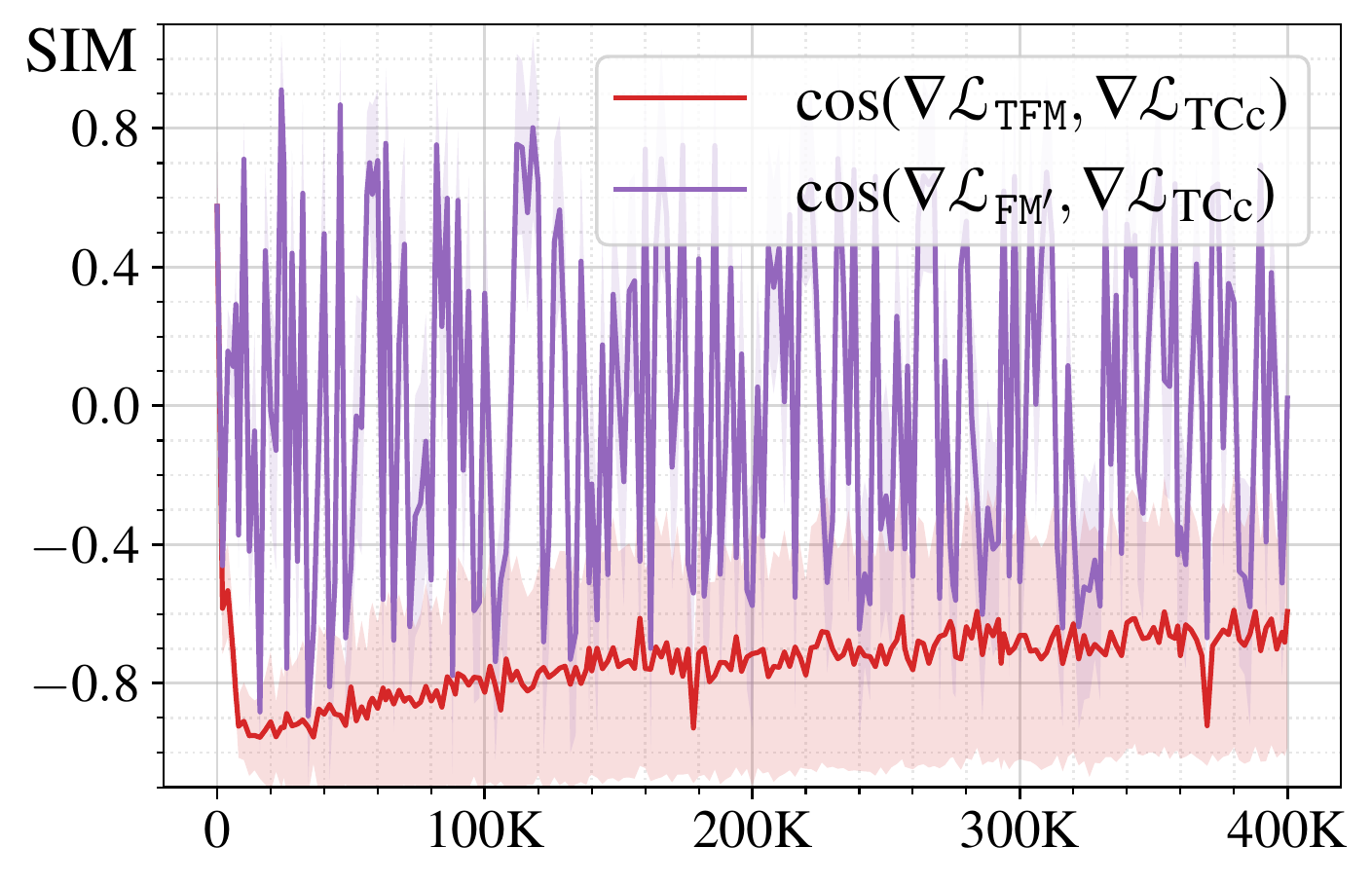}
\caption{Gradient similarity}
\label{fig:grad-sim}
\end{subfigure}
\begin{subfigure}{0.32\textwidth}
\includegraphics[width=\linewidth]{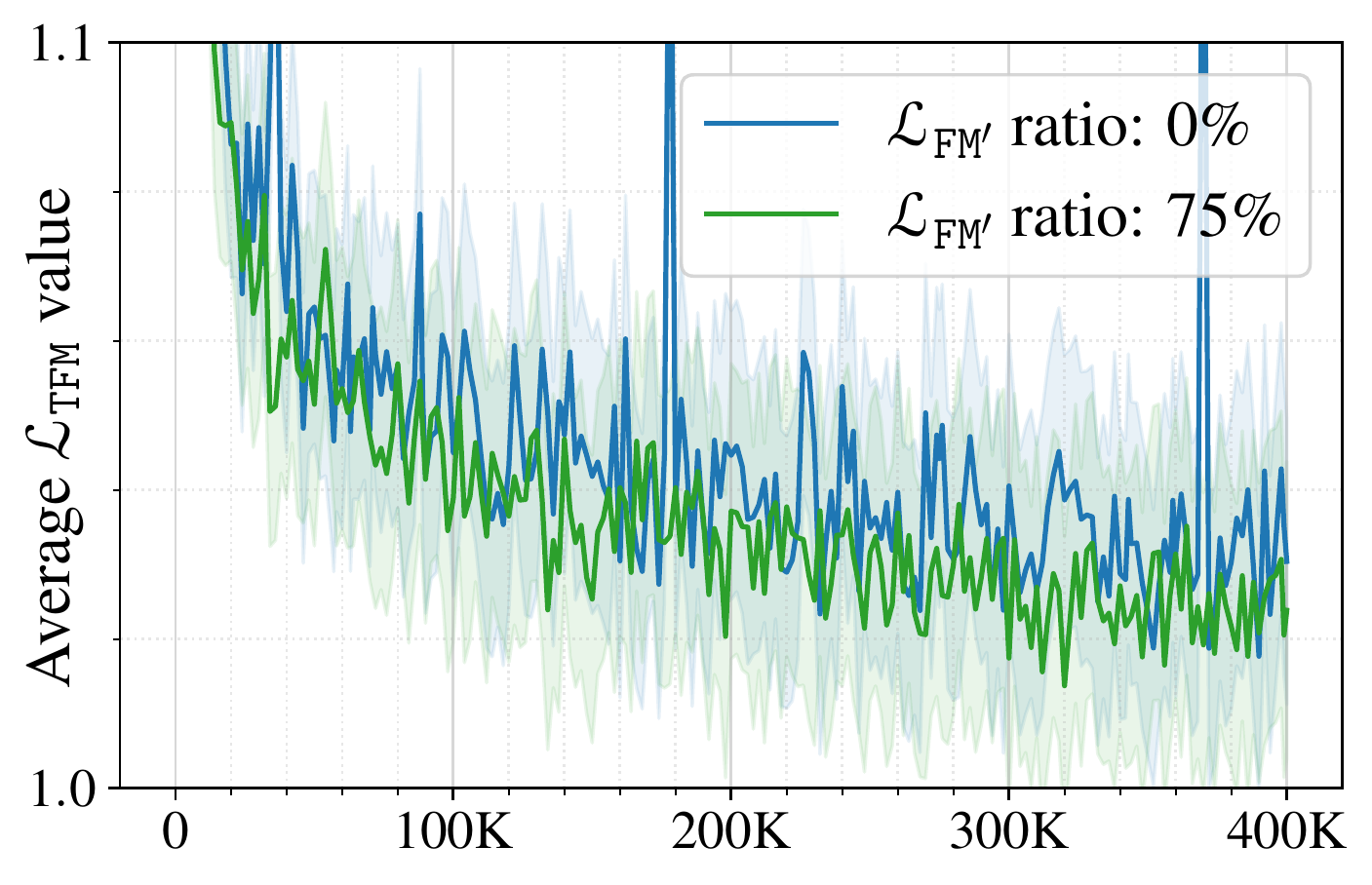}
\caption{$\LuFM$ under different $\LbFM$ ratios.}
\label{fig:ufm_per_bfm}
\end{subfigure}
\begin{subfigure}{0.32\textwidth}
\includegraphics[width=\linewidth]{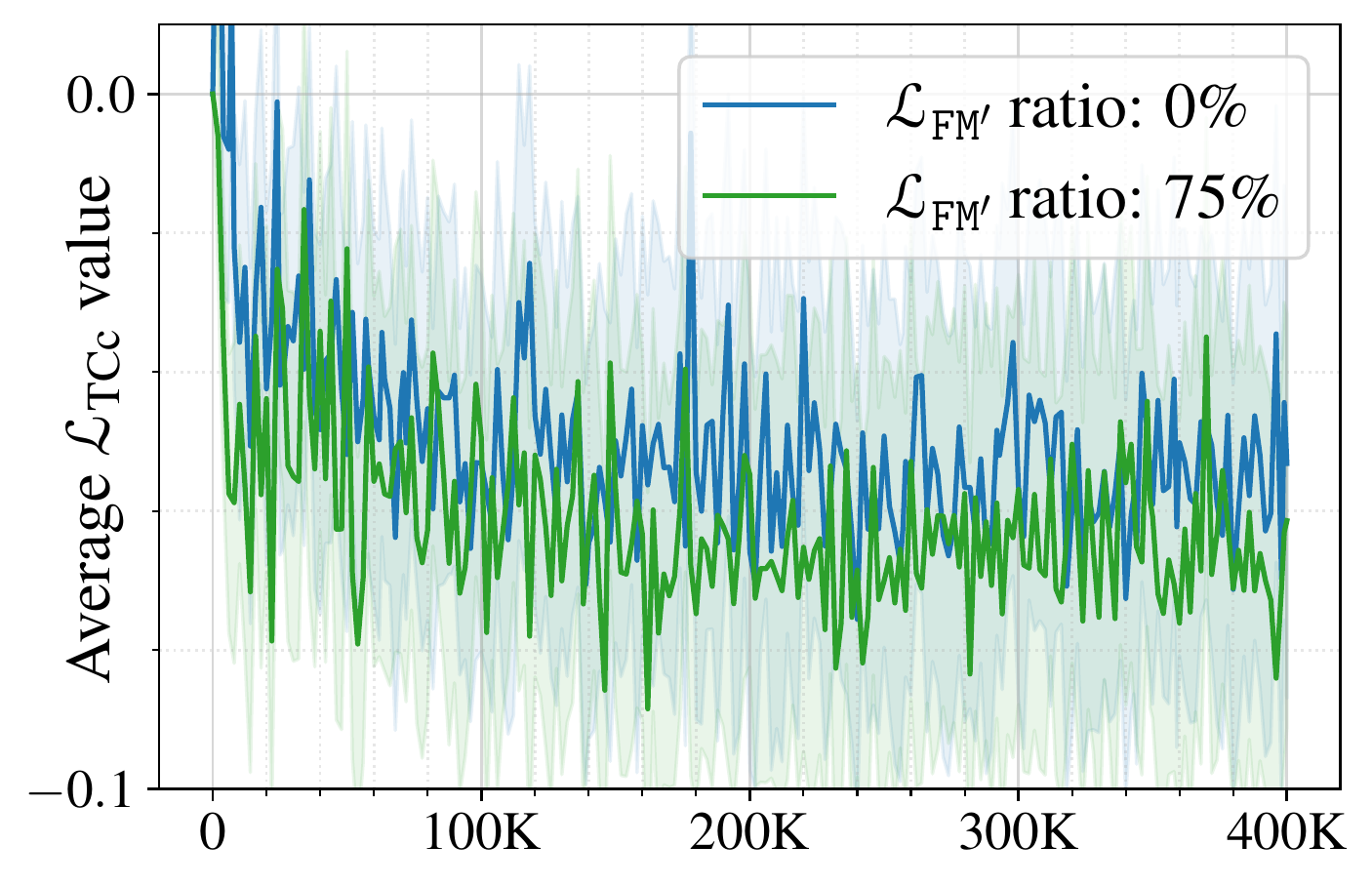}
\caption{$\LCFc$ under different $\LbFM$ ratios.}
\label{fig:cfc_per_bfm}
\end{subfigure}
\caption{\textbf{MeanFlow training analysis}. (a) Shows the cosine similarity between the gradients of two loss pairs ($\nabla \LCFc$ vs. $\nabla \LuFM$ and $\nabla \LCFc$ vs. $\nabla \LbFM$) throughout training. (b) Evaluated $\LuFM$ when MeanFlow trained with 0\% and 75\% of $\LbFM$. (c) Evaluated  $\LCFc$ when MeanFlow trained with 0\% and 75\% of $\LbFM$.}
\label{fig:analysis}
\end{figure}


An intriguing aspect of MeanFlow is the noise distribution used during training: \cite{MeanFlow} empirically found that the best results are achieved when setting $r = t$ for 75\% of the samples.
This might look counter-intuitive, since we are interested in learning the average velocity on a $[r, t]$ interval to perform large trajectory leaps during inference, so why spending the majority of the training computation on fitting this border case that corresponds to vanilla flow matching supervision?
In this section, we show that the MeanFlow loss on its own can be interpreted as velocity consistency training with extra flow matching supervision, and analyze the interaction of these two objectives.

\subsection{Understanding the objective}
\label{sec:analysis:loss}

Through algebraic manipulations, the original MeanFlow loss $\LMFc$ in Eq.~\eqref{eq:mean_flow_loss} can be rewritten into the following equivalent form (see \Cref{app:proof_decomp}):
\begin{equation}
\begin{aligned}
\label{eq:lmf-decomposed}
\LMFc(\param) = \underbrace{\expecttrzt{\|\vu_{\param}(\vz_t, r, t) - \vvt \|^2_2}}_\text{\UFMname \LuFM} + \underbrace{\expecttrzt{2\left(t - r\right) \cdot \vu_{\param}^{\top}(\vz_t, r, t) \frac{\mathrm{d} \vu_{\paramsg}(\vz_t, r, t)}{\mathrm{d} t}}}_\text{\CFname \LCFc} + C,\\
\end{aligned}
\end{equation}
where $C$ is a constant independent of $\param$.
In this decomposition, the first term \LuFM corresponds to a flow matching loss but with an additional modeling input parameter $r$, so we refer to it as \emph{\uFMname}.
The second term \LCFc, denoted as \emph{\cFname} loss, acts as a $(t-r)$-reweighted continuous consistency loss
\footnote{Similarly to the proof in Remark 10 of \cite{CMs}, one can show that this term is equivalent to minimizing the difference between $\vu_{\param}(\vz_t, r, t)$ and $\vu_{\paramsg}(\vz_{t-\Delta t}, r, t-\Delta t)$ as $\Delta t \rightarrow 0$.},
but also without a boundary condition~\citep{CMs}.
This decomposition highlights that the MeanFlow objective can be interpreted as a consistency (trajectory) model with extra flow matching supervision.

An interesting property of this decomposition is that \LCFc does not have any boundary condition.  
In comparison, \cite{CMs} enforces such a condition for vanilla consistency models using a $\vzzero$-prediction parameterization: without it, the model would quickly converge to a trivial solution (e.g., a constant output).
In the MeanFlow case, this collapse does not occur, which suggests that \LuFM implicitly provides the boundary condition for \LCFc.
We believe that the absence of an explicit boundary condition makes \LCFc easier to optimize and gives it a much larger solution space.  

Another important observation here is that \uFMname involves random $r \leqslant t$, which differs from the $r = t$ case used during training by \cite{MeanFlow}.
To clarify this distinction, we directly compare \uFMname (\LuFM) with vanilla flow matching, which we denote as \LbFM when using the $u$-prediction parameterization:
\begin{equation}
\LuFM \triangleq \expect[t,r,\vzt]{\|\vu_{\param}(\vz_t, r, t) - \vvt \|^2_2},
\qquad
\LbFM \triangleq \expect[t,r,\vzt \textcolor{azure}{|r=t}]{\|\vu_{\param}(\vz_t, r, t) - \vvt \|^2_2}
\end{equation}

Here, \LuFM arises from the decomposition of the MeanFlow loss, while \LbFM corresponds to the objective used in \cite{MeanFlow} for joint training.
From this formulation, several observations follow.
First, \LbFM is a ``part'' of \LuFM, active only on the $p(t,r \mid r=t)$ slice of the joint distribution $p(t,r)$.
Second, if the network is independent of $r$, then marginalizing out $r$ yields $\LuFM \equiv \LbFM$, reducing the objective to vanilla flow matching.



\subsection{Empirical Analysis}
\label{sec:analysis:grads}
With the decomposition in \Cref{eq:lmf-decomposed}, how does $\LbFM$ interact with the two decomposed terms?
In this section, we analyze the gradients of these losses and examine how extra $\LbFM$ minimization affects $\LuFM$ and $\LCFc$ individually.
We conduct detailed experiments by training MeanFlow with the DiT-B/2 \citep{DiT} architecture on \INFull \citep{imagenet} for 400K steps. 
Additional experiment settings are in \Cref{app:analysis-details}.

We first analyze the training dynamics by measuring the cosine similarity between the gradients $\nabla\LuFM$ and $\nabla\LCFc$ during training.
As shown in \Cref{fig:grad-sim}, these two gradients are strongly negatively correlated, with a similarity typically below $-0.4$.
This reveals that optimizing $\LuFM$ and $\LCFc$ jointly is inherently difficult. 
We hypothesize this stems from the fact that $\LCFc$, without any boundary condition, has a very large optimal solution manifold, compared to $\LuFM$ whose manifold is very narrow.
Thus the optimization process is getting pulled towards the $\LCFc$ manifold, distracting from reaching a narrow intersection.

Given this gradient conflict, the question arises: why does joint training with $\LbFM$ help?
We identify two key reasons: First, as a subset of $\LuFM$, $\LbFM$ directly reduces $\LuFM$.
This is empirically confirmed in \Cref{fig:ufm_per_bfm}, where allocating 75\% of the training budget to $\LbFM$ significantly lowers the overall $\LuFM$ compared to pure MeanFlow training.
Second, $\LbFM$ applies only at $r=t$, where $\LCFc = 0$. 
Consequently, the gradient $\nabla\LbFM$ interferes less with $\nabla\LCFc$ than the $\nabla\LuFM$ gradient.
This is demonstrated in \Cref{fig:grad-sim}, which shows that $\cos(\nabla\LbFM,\nabla\LCFc)$ is consistently higher than $\cos(\nabla\LuFM,\nabla\LCFc)$, that is strongly negative for more than 95\% of the training.
Surprisingly, $\LCFc$ component doesn't seem to be affected and can even be lower when allocating 75\% of the training budget to $\LbFM$, as shown in \Cref{fig:cfc_per_bfm}.
Which again hints at the fact that $\LCFc$ is relatively easy to optimize, even near the $\LuFM$ optimum. 

In conclusion, our analysis reveals three important observations:
\textit{
\begin{tcolorbox}
\begin{center}
\begin{itemize}[leftmargin=*]
\item[$\triangleright$] \LMFc can be decomposed into \uFMname \LuFM and \cFname \LCFc objectives, whose gradients are strongly negatively correlated during training.
\item[$\triangleright$] \LCFc does not have a necessary boundary condition on its own, implying that \LuFM serves as an implicit boundary condition for it.
\item[$\triangleright$] \LbFM acts as a surrogate loss for \LuFM, but with significantly less gradient conflict with the \CFname loss \LCFc. 
\end{itemize}
\end{center}
\end{tcolorbox}
}

\section[AlphaFlow Models]{\modelname models}
\label{sec:alphaflow}

As we showed in the previous section, the $\LuFM$ loss is difficult to optimize jointly with the $\LCFc$.
While the introduction of the $\LbFM$ loss serves as an effective surrogate for optimizing $\LuFM$, this approach dedicates a significant portion of training to an objective that is not of our primary interest.
This raises a key question: \textit{Can we more efficiently optimize $\LuFM$ when optimizing $\LMFc$ without this computational overhead?}
To answer this, we introduce our \modelname loss, a new family of training objectives for flow-based models.


\subsection[AlphaFlow: Unifying one, few, and many-step flow-based models]{\modelname: Unifying one, few, and many-step flow-based models}
\label{sec:af:unify}

\begin{defi}
The \modelname loss $\LMFd$ is defined as:
\begin{equation}
\begin{aligned}
\label{eq:discrete-meanflow-loss}
\LMFd(\param) &\triangleq \expecttrzt{\stepratio^{-1} \cdot \left\Vert\vutheta(\vz_t, r, t) - \left(\stepratio \cdot \vvoffset +  (1 - \stepratio) \cdot \vuthetasg(\vz_s, r, s)\right) \right\Vert^2_2 },
\end{aligned}
\end{equation}
where $t, r \in [0, 1]$ is the start and end timestep, $s$ is the \sname: $s = \stepratio \cdot r + (1 - \stepratio) \cdot t, \stepratio \in (0, 1]$ is the \rationame, and 
$\vz_s = \vz_{t} + (t - s) \cdot \vvoffset$ is the trajectory value at this timestep $s$.
Here, $\vvoffset$ is the ``shift velocity'' used to estimate the intermediate variable $\vzs$ from $\vzt$.
\end{defi}

\begin{figure}
\centering
\begin{subfigure}{0.19\textwidth}
    \includegraphics[height=1.5cm]{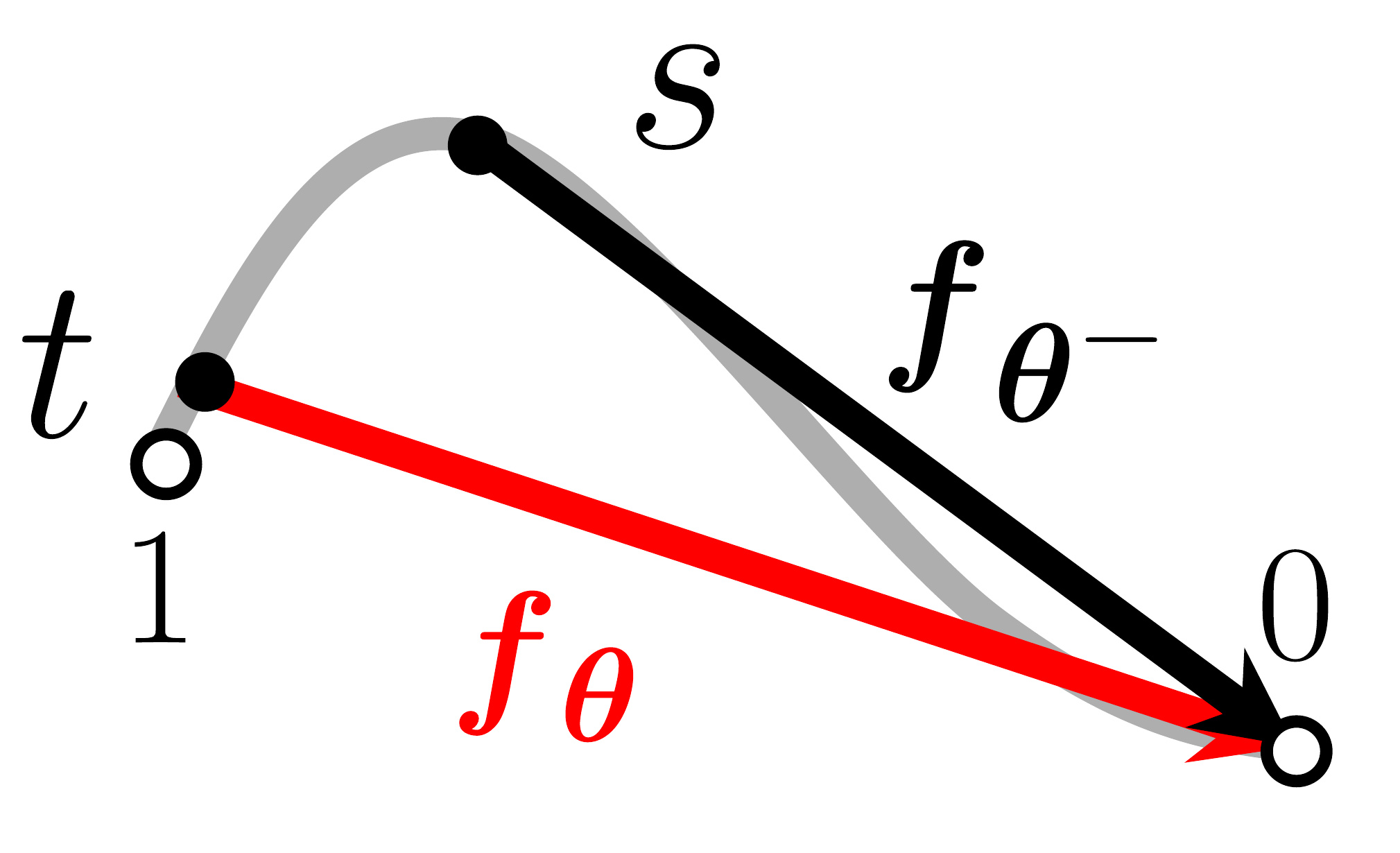}
    \caption{Discrete CT}
    \label{fig:CTd}
\end{subfigure}
\hfill
\begin{subfigure}{0.19\textwidth}
    \includegraphics[height=1.5cm]{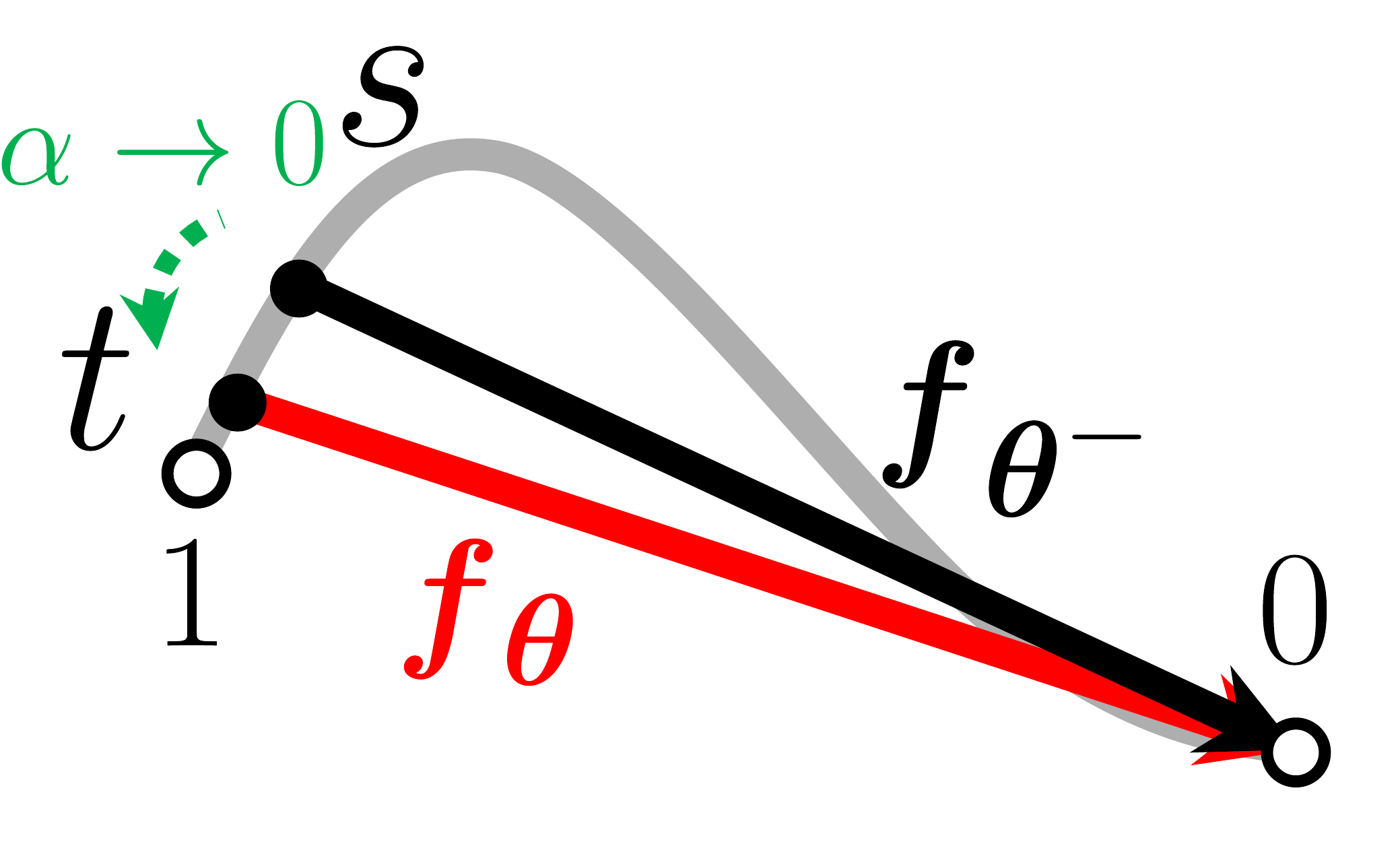}
    \caption{Continuous CT}
    \label{fig:CTc}
\end{subfigure}
\hfill
\begin{subfigure}{0.19\textwidth}
    \includegraphics[height=1.5cm]{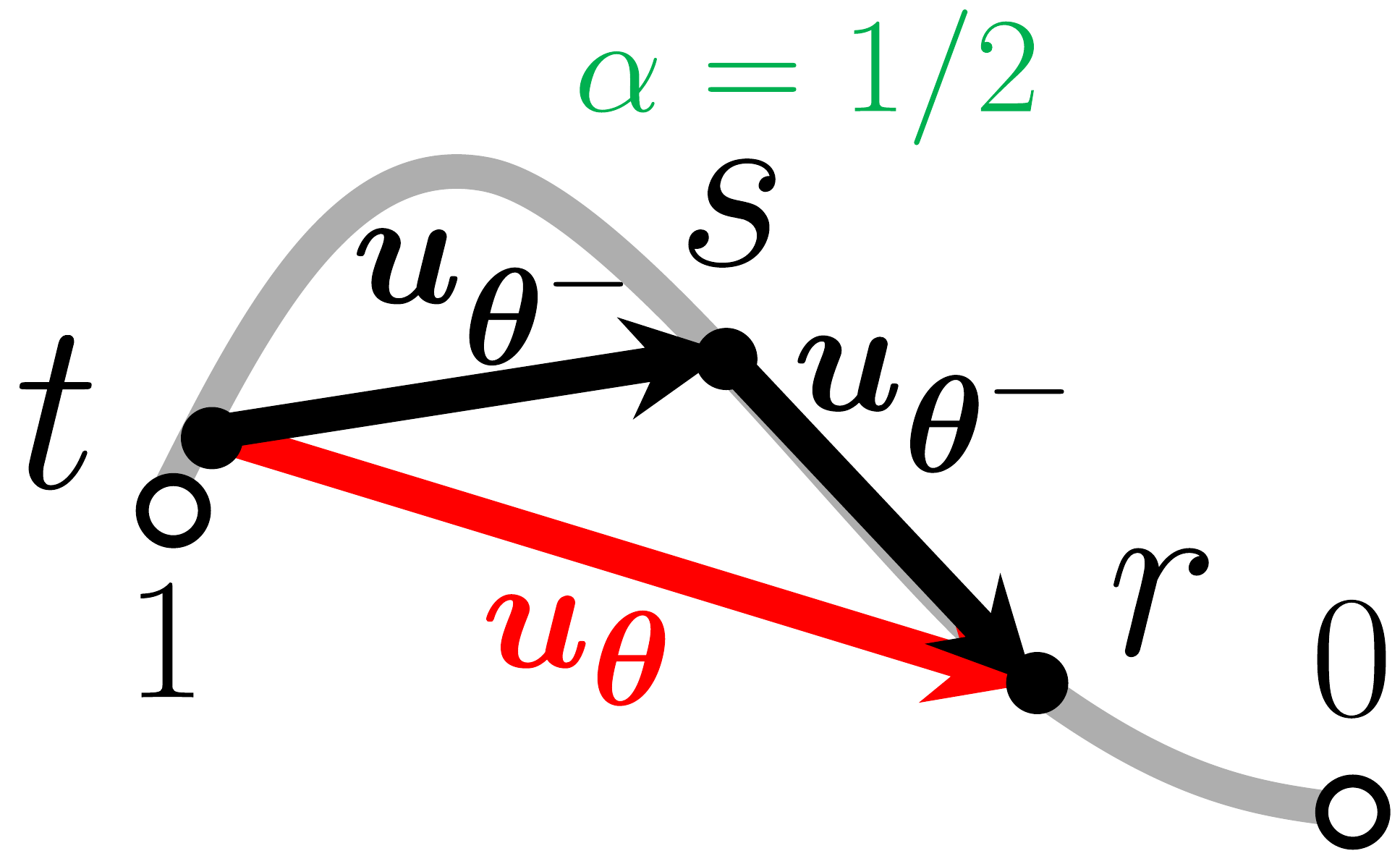}
    \caption{Shortcut Model}
    \label{fig:SC}
\end{subfigure}
\hfill
\begin{subfigure}{0.19\textwidth}
    \includegraphics[height=1.5cm]{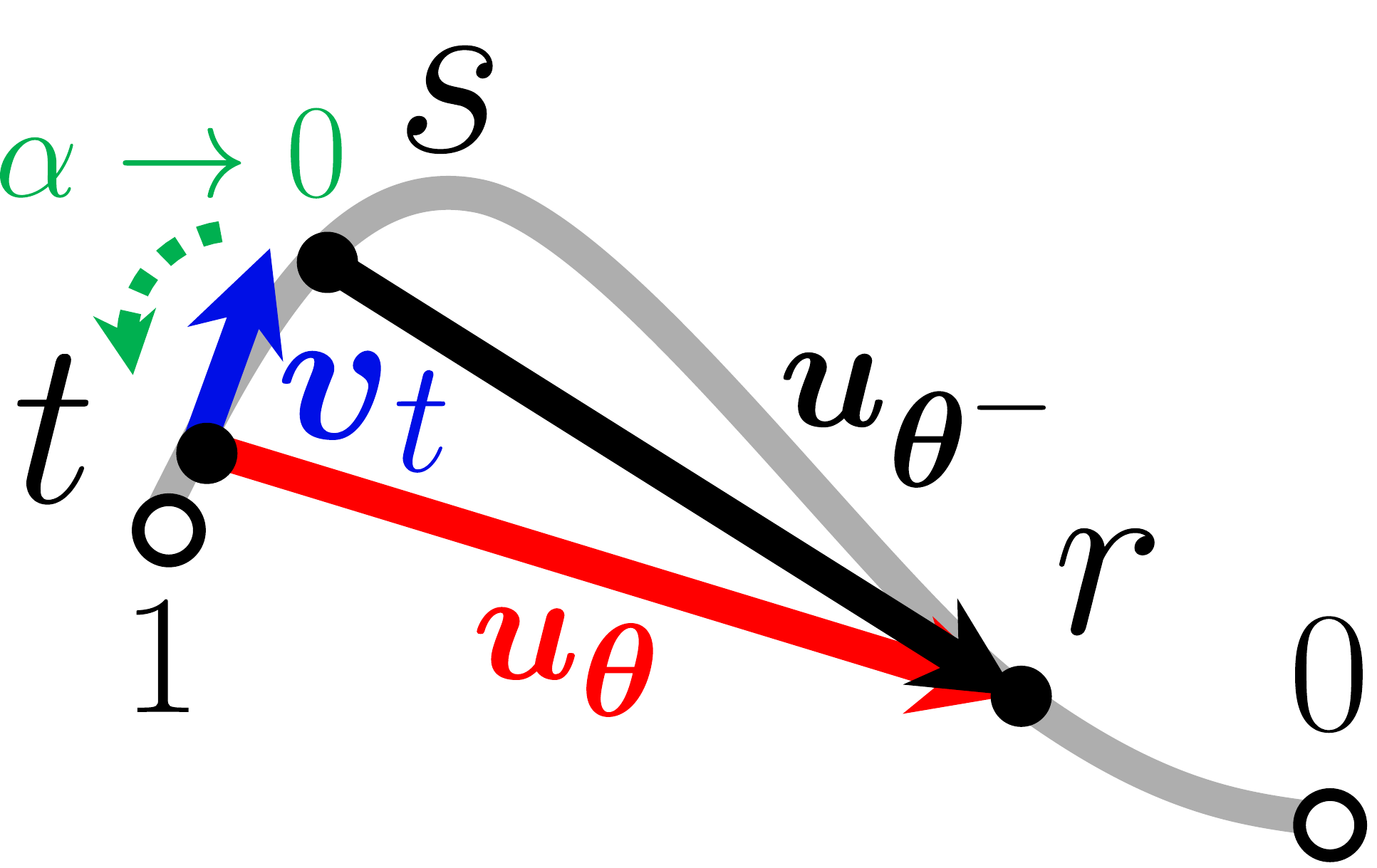}
    \caption{MeanFlow}
    \label{fig:MF}
\end{subfigure}
\hfill
\begin{subfigure}{0.19\textwidth}
    \includegraphics[height=1.5cm]{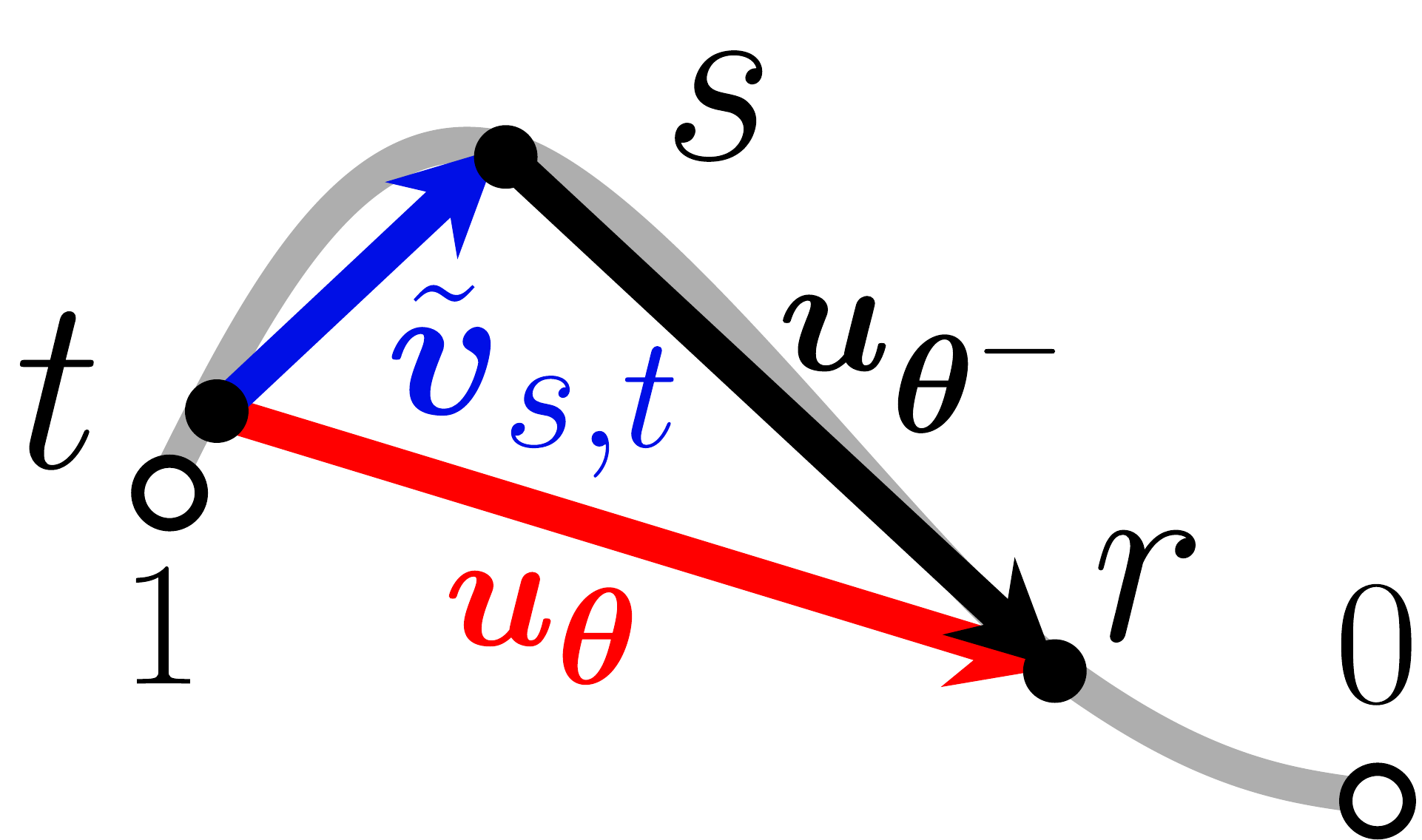}
    \caption{\modelname}
    \label{fig:AF}
\end{subfigure}
\caption{Comparison of training trajectories for various few-step diffusion and flow-based models.}
\end{figure}

The \modelname loss is visualized in \Cref{fig:AF}.
Intuitively, it enforces trajectory consistency between $t$ and $r$ by introducing an additional $s$, which is an interpolation between $t, r$ with ratio $\stepratio$. 
More importantly, this definition generalizes previously introduced training objectives such as \uFMname, Shortcut Model training, and MeanFlow training:

\begin{thm}
\label{thm:unified-loss}
The \modelname loss unifies flow matching, Shortcut Models, and MeanFlow:
\begin{itemize}[leftmargin=*]
\item[$\triangleright$] $\LuFM(\param) = \mathcal{L}_{\alpha = 1}(\param)$ with $\vvoffset = \vvt$.
\item[$\triangleright$] $\LSC(\param) = \frac{1}{2} \mathcal{L}_{\alpha = 1/2}(\param)$ with $\vvoffset = \vu_{\param^{-}}(\vz_t, s, t)$.
\item[$\triangleright$] $\nabla_{\param}\LMFc(\param) = \nabla_{\param}\mathcal{L}_{\alpha \to 0}(\param)$ with $\vvoffset = \vvt$.
\end{itemize}
\end{thm}

Moreover, if one considers a $\vzzero$-parametrized network $\vu_{\param}(\vz_t, 0, t) = \left(\vz_t - \vf_{\param}(\vz_t, t)\right)/t = \vzzerohat$, $\LMFd$ incorporates discrete and continuous consistency training as well.
Specifically, with $\vvoffset = \vvt$ and $r \equiv 0$:

With $\vu_{\param}(\vz_t, 0, t) = \left(\vz_t - \vf_{\param}(\vz_t, t)\right)/t = \vzzerohat$, $\vvoffset = \vvt$ and $r \equiv 0$:

\begin{itemize}[leftmargin=*]
\item[$\triangleright$] $\LCTd(\param) = \mathcal{L}_{\alpha = \delta}(\param)$ for $\delta \in (0, t)$.
\item[$\triangleright$] $\nabla_{\param}\LCTc(\param) = \nabla_{\param}\mathcal{L}_{\alpha \to 0}(\param)$.
\end{itemize}

\begin{wrapfigure}{r}{.6\linewidth} 
\centering
\vspace{-.9in}
\resizebox{\linewidth}{!}{%
\begin{minipage}{\linewidth}
\input{algos/training}
\vspace{-0.2in}
\input{algos/schedule}
\end{minipage}
}
\vspace{-0.3in}
\end{wrapfigure}


This theorem reveals that the ratio $\stepratio$ is the key hyperparameter that unifies seemingly different methods, which controls the relative position of the intermediate timestep $s$ within the $(r, t)$ interval. 
By annealing $\stepratio$ from 1 to 0, we obtain a family of models in the interpolation between \uFMname and MeanFlow. 
Notably, discrete CT is a special case of \modelname with $r \equiv 0$. 
Unlike discrete CT, \modelname requires no complex timestep partitioning: once $t$ and $r$ are sampled, $s$ is immediately determined with a fixed $\stepratio$.





\subsection{\texorpdfstring{\modelname}{AlphaFlow} models}

The \modelname loss enables a curriculum learning strategy that progressively transitions from the \uFMname to MeanFlow objective. 
This approach better disentangles the optimization of the \uFMname and consistency losses, could potentially reduce reliance on the \bFMname objective, and leads to better convergence.
The detailed curriculum learning can be summarized into three phases:
\begin{itemize}[leftmargin=*]
\item \textbf{\UFMname pretraining} ($\alpha = 1$). 
To speed-up convergence toward narrow $\LuFM$ manifold, we prioritize optimizing \uFMname in the early training phase. 
Additionally, as a low-variance objective, \uFMname quickly establishes a reliable noise-to-data mapping, providing a good initialization for subsequent few-step refinement. 
Notably, this pretraining strategy is aligned with previous diffusion model pretraining strategy applied on consistency model \citep{EasyCM}, while we start from different motivations and generalize it into the \modelname framework.
\item \textbf{\modelname transition} ($\alpha \in (0, 1)$). Once the model has a solid foundation, we transition from \uFMname to the MeanFlow objective. 
We accomplish this with a curriculum learning approach where we progressively decrease the \rationame from 1 to 0. 
This gradual shift is inspired by the discrete CT \citep{CMs}. 
It effectively transitions from the ``high-bias, low-variance'' objective to the ``high-variance, low-bias one'', leading to improved convergence.
\item \textbf{MeanFlow fine-tuning} ($\alpha \rightarrow 0$). 
In the final stage, we focus entirely on the MeanFlow training objective. Unlike the original paper, our improved early-stage optimization of \uFMname significantly reduces the need for the \bFMname loss (as shown in \Cref{tab:ablation_alphaflow} (b)) and achieves significantly better few-step generation quality.
\end{itemize}
The overall training code of \modelname is shown in \Cref{alg:code}, where we first sample $t, r$ and obtain the $\stepratio$ from the schedule. 
Based on whether $\stepratio = 0$ or not, \modelname will use either $\LMFc$ or $\LMFd$ to train the model. 
\modelname applies the same training details as MeanFlow when training $\LMFc$ (except a lower ratio of flow matching). 
Below, we only show the difference: the schedule of $\stepratio$ as well as the design space of $\LMFd$ when $\alpha > 0$.
\vspace{-0.1in}
\inlinesection{Schedule} To schedule the training, we use a sigmoid function, $\stepratio = \mathtt{Sigmoid}_{k_s \Rightarrow k_e, \gamma, \eta}\left(k\right)$, which depends on the training iteration $k$.
The function is defined by its starting and ending iterations, $k_s, k_e$, a temperature parameter $\gamma$ (set to be 25) and a clamping value $\eta$. 
The specific implementation can be found in  \Cref{alg:code_schedule}. 
\Cref{fig:schedule_vis} provides a visualization of this scheduler, while \Cref{sec:ablation} conducts an ablation study over its parameters.

\vspace{-0.2in}

\inlinesection{Clamping value}
\cite{EasyCM} show that when $\Delta t = t - s$ approaches 0, the performance of few-step CT model will first increase and then decrease. 
For \modelname, we observe a similar phenomenon: by training \modelname with a fixed $\stepratio$, as $\stepratio$ approaches 0, the 1-step generation performance will first increase then decrease. 
Detailed experiments are shown in \Cref{tab:discrete_ablation} (c). From the experiment, the optimal performance is achieved when $\stepratio = 5 \times 10^{-3}$. 
Thus, we set a clamping value $\eta = 5 \times 10^{-3}$ for the schedule. $\stepratio$ will be set to $0$ when $\stepratio < \eta$. 
We also use the same clamping value to set $\stepratio$ to $1$ when $\stepratio > 1 - \eta$, as when $\stepratio$ is close to $1$, $\LuFM$ is similar to $\LMFd$ but more efficient.

\vspace{-0.2in}
\inlinesection{Training objective}
In the unifying space of \modelname loss, all other few-step models set $\vvoffset = \vv_t$ except the shortcut model which uses $\vvoffset = \vu_{\param^{-}}(\vz_t, s, t)$. 
Additionally, we are interested in seeing whether we need exponential moving average (EMA) for $\paramsg$. 
With ablation study in \Cref{tab:discrete_ablation} (a), we set $\vvoffset = \vv_t$ and do not use EMA for $\paramsg$. 
\vspace{-0.1in}
\inlinesection{Adaptive loss weight} MeanFlow \citep{MeanFlow} demonstrates the effectiveness of adaptive loss. 
Basically, let $||\Delta||_2^2$ denote the squared L2 loss. 
The adaptive loss weight $\omega = 1/(||\Delta||_2^2 + c)$ where $c = 10^{-3}$. 
And the adaptively weighted loss is $\mathtt{sg}(\omega)||\Delta||_2^2$. 
Theoretically, we derived an equivalent adaptive loss weight $\omega = \stepratio/(||\Delta||_2^2 + c)$ for $\LMFd$. 
We defer the derivation in \Cref{app:ablation-discrete-meanflow}. 
With ablation study in \Cref{tab:discrete_ablation} (b), we demonstrate the derived adaptive loss weight is better than other loss weights.
\vspace{-0.1in}
\inlinesection{Classifier-free guidance (CFG)}
We apply a similar CFG training strategy as MeanFlow, by setting $\vvoffset$ in \Cref{eq:discrete-meanflow-loss} with $\vvoffset = \omegacfg \cdot \vv(\vz_t, t|\vx) + \kappacfg \cdot \vuthetasg\left(\vz_t, t, t|\bm c\right) + \left(1 - \omegacfg - \kappacfg\right) \cdot \vuthetasg\left(\vz_t, t, t|\varnothing\right)$, where $\omegacfg, \kappacfg$ are the guidance scale, $\vuthetasg\left(\cdot|\bm c\right)$, $\vuthetasg\left(\cdot|\varnothing\right)$ denotes the class-condition (with class $c$) and class-unconditional prediction.
Detailed settings of $\omegacfg, \kappacfg$ are deferred to \Cref{sec:impl-details}.
\vspace{-0.1in}
\inlinesection{Sampling} We employ both consistency sampling \citep{CMs} and ODE sampling for two-step generation. Implementation details are provided in \Cref{alg:code_sample}. Empirically, we observe that consistency sampling outperforms ODE sampling for larger models with better convergence. Consequently, we adopt ODE sampling for all DiT-B/2 architectures and consistency sampling for all DiT-XL/2 architectures, with additional ablation studies on DiT-XL/2 presented in \Cref{fig:sampling}.

\section{Experiments}\label{sec:experiments}

\newcommand{\noclsbalancedres}{\cellcolor{gray!30}\emptytable}
\begin{table*}[t]
\centering
\tablestyle{2pt}{1.}
\small
\begin{tabular}{llcc ccccc}
\toprule
\multirow{2}{*}{Method} & \multirow{2}{*}{Source} & \multirow{2}{*}{Params} & \multirow{2}{*}{Epochs} & \multicolumn{2}{c}{NFE 1} & \multicolumn{3}{c}{NFE 2} \\
\cmidrule(lr){5-6} \cmidrule(lr){7-9}
& & & & \FID & \FDD & \FID & \FDD & $\text{\FID}^{\dag}$ \\
\midrule
Shortcut-XL/2 & \cite{ShortcutModels} & 675M & 160 & 10.60 & \emptytable & \emptytable & \emptytable & \noclsbalancedres\\
IMM-XL/2 & \cite{IMM} & 676M & 3840 & 8.05 & \emptytable & 3.88 & \emptytable & \noclsbalancedres\\
MeanFlow-XL/2 & \cite{MeanFlow} & 676M & 240 & 3.43 & \emptytable & 2.93 & \emptytable & \noclsbalancedres \\
MeanFlow-XL/2+ & \cite{MeanFlow} & 676M & 1000 & \emptytable & \emptytable & 2.20 & \emptytable & \noclsbalancedres \\
FACM-XL/2 & \cite{FACM} 
& 675M  & 800 + $250 \times 2$ & \emptytable & \emptytable & \emptytable & \emptytable & \cellcolor{gray!30} 2.07 \\
\midrule
FACM-XL/2 & \multirow{4}{*}{Our reproduction}
& 675M  & $120 \times 2$ & 9.54 & 410.4 & 7.31 & 362.0 & \noclsbalancedres \\
FACM-XL/2 & & 675M & $240 \times 2$ & 6.59 & 327.7 & 4.73 & 278.6  & \noclsbalancedres \\
MeanFlow-B/2 & & 131M & 240 & 6.04 & 312.3 & 5.17 & 232.1 & \noclsbalancedres \\
MeanFlow-XL/2 & & 676M & 240 & 3.47 & 185.8 & 2.46 & 108.7 & \cellcolor{gray!30} 2.26 \\
\midrule
\modelname-B/2 & \multirow{3}{*}{Our methods} & 131M & 240 & 5.40 & 287.1 & 5.01 & 231.8 & \noclsbalancedres \\
\modelname-XL/2 & & 676M & 240 & 2.95 & 164.6 & 2.34 & 105.7 & \cellcolor{gray!30} 2.16\\
\modelnamexlp & & 676M & 240+60 & \textbf{2.58} & \textbf{148.4} & $\textbf{2.15}$ & \textbf{96.8} & \cellcolor{gray!30}\textbf{1.95}\\
\bottomrule
\end{tabular}

\caption{\textbf{Class-conditional generation on ImageNet-256$\times$256}. 
The table reports the results for few-step diffusion/flow matching-based methods trained from scratch. 
"$\times$2" indicates that FACM requires roughly twice the computation per epoch compared to other methods. 
For a direct "epoch-to-epoch comparison," \modelname-XL/2, MeanFlow-XL/2 and FACM-XL/2 are each trained for 240 epochs.
\modelnamexlp is a fine-tuned version of \modelname-XL/2, trained for extra 60 epochs with a batch size of 1024.
$\dag$ FID scores are evaluated with the balanced class sampling (see \Cref{app:random_balance_class}). 
}
\label{tab:imagenet256}
\end{table*}

In this section, we employ \modelname on real image datasets \INFull \cite{imagenet}.
We use exactly the same DiT~\cite{DiT} architecture as MeanFlow~\cite{MeanFlow}.
For evaluation, we use Fréchet Inception Distance (\FID) \cite{FID}, Fréchet DINOv2~\cite{DINOv2}.
We evaluate model performance for both 1 and 2 Number of Function Evaluations (NFE=1, NFE=2).
We implement our models in the latent space of the Stable Diffusion Variational Autoencoder (SD-VAE) \footnote{The EMA version in \url{https://huggingface.co/stabilityai/sd-vae-ft-mse}}.
More details on the experiments settings are in \Cref{sec:impl-details}.
\vspace{-0.1in}
\subsection{Comparison with baseline}
\label{sec:comparion_baseline}

In \Cref{tab:imagenet256}, we compare \modelname with previous few-step Diffusion and Flow models, demonstrating its superior performance for 1-NFE and 2-NFE generation.
Across models trained for 240 epochs, \modelname-XL/2 achieves \textbf{2.95} FID (\textbf{164.6} FDD), representing a relative improvement of 15\% (12\%) over MeanFlow-XL/2 and 70\% (60\%) over FACM-XL/2. 
Our best model, \modelname-XL/2+, sets a new state-of-the-art 1-NFE generation with an impressive FID of \textbf{2.58} (\textbf{148.4} FDD), compared with all the other few-step Diffusion and Flow models trained over the SD-VAE.
Furthermore, for 2-NFE generation, \modelname-XL/2+ achieves \textbf{2.15} FID (\textbf{96.8} FDD), outperforms all these baseline methods.
It's particularly notable that it surpasses FACM-XL/2's 2.07 FID (achieved with a class-balanced sampling) by reaching 1.95 FID with only 23\% of the training epochs. 
Uncurated samples, shown in \Cref{fig:teaser} and \Cref{app:additional_vis}, visually confirm these results. Specifically in \Cref{fig:teaser}, \modelname-XL/2 generates more images with better quality, as highlighted in green.




\subsection{Ablation Study}

\label{sec:ablation}
\inlinesection{\Rationame schedule}
\begin{table}[t]
\centering
\begin{minipage}{.45\textwidth}
    \centering
    \tablestyle{2pt}{.9}
    \small
    \begin{tabular}{l cccc}
        \toprule
        \multirow{2}{*}{Schedule}  & \multicolumn{2}{c}{NFE 1} & \multicolumn{2}{c}{NFE 2} \\
        \cmidrule(lr){2-3} \cmidrule(lr){4-5}
        & \FID & \FDD & \FID & \FDD \\
        \midrule
         $\mathtt{Constant}_{0.0}$ & 44.4 & 844.1 & 42.1 & 836.3 \\
         \midrule
         \multicolumn{5}{l}{\textit{\UFMname iterations}} \\
        $\SigmoidSch{0}{100}$ &  44.3 & 860.3 & 40.8 & 826.9 \\
        $\SigmoidSch{50}{150}$  & 44.1 & 846.8 & 39.9 & 811.6\\
         $\SigmoidSch{100}{200}$ & 42.4 & 828.0 & 38.3 & 795.4\\
         $\SigmoidSch{150}{250}$ & 41.3 & 818.8 & 38.1 & 793.1 \\
        \midrule
        \multicolumn{5}{l}{\textit{Transition iterations}} \\
        $\SigmoidSch{200}{200}$ & 41.4 & 794.4 & 38.8 & 796.7 \\
        $\SigmoidSch{150}{250}$ & 41.3 & 818.8 & 38.1 & 793.1 \\
        $\SigmoidSch{0}{400}$  & \textbf{40.0} & \textbf{785.4} & \textbf{37.1} & \textbf{782.9} \\       
        \bottomrule
    \end{tabular}
    \caption*{(a) \textbf{\Rationame schedule.}}
\end{minipage}%
\begin{minipage}{.53\textwidth}
\centering
\tablestyle{2pt}{1.0}
\small
\begin{tabular}{cl cccc}
\toprule
\multicolumn{2}{c}{Model} & \multicolumn{2}{c}{NFE 1} & \multicolumn{2}{c}{NFE 2} \\
\cmidrule(lr){1-2} \cmidrule(lr){3-4} \cmidrule(lr){5-6}
\% $r=t$ & Schedule & \FID & \FDD & \FID & \FDD \\
\midrule
 \multirow{2}{*}{0\%} & $\mathtt{Constant}_{0.0}$ & 46.0 & 879.6 & 44.3 & 867.7 \\
& $\SigmoidSch{0}{400}$ & 40.4  & 822.5 & 38.9 & 811.8 \\
\midrule
 \multirow{2}{*}{25\%} & $\mathtt{Constant}_{0.0}$ & 44.4 & 844.1 & 42.1 & 836.3 \\
& $\SigmoidSch{0}{400}$ & \textbf{40.0}  & 785.4 & 37.1 & 782.9 \\
\midrule
 \multirow{2}{*}{50\%} & $\mathtt{Constant}_{0.0}$ & 43.9 & 844.1 & 42.1 & 836.3 \\
& $\SigmoidSch{0}{400}$ & 40.2  & \textbf{781.0}  & 37.1  & 775.0 \\
 \midrule
 \multirow{2}{*}{75\%} & $\mathtt{Constant}_{0.0}$ & 43.1 & 819.2 & 38.5 & 787.6 \\
& $\SigmoidSch{0}{400}$ & 42.2  & 810.5 & \textbf{36.2} & \textbf{754.7} \\   
\bottomrule
\end{tabular}
\caption*{(b) \textbf{Flow matching ratio.}}
\end{minipage}
\caption{Ablation study on \INFull for \modelname-B/2.}
\label{tab:ablation_alphaflow}
\vspace{-0.2in}
\end{table}

\begin{figure}[t]
\centering
\begin{minipage}{.48\textwidth}
   \centering
  \includegraphics[height=4.2cm]{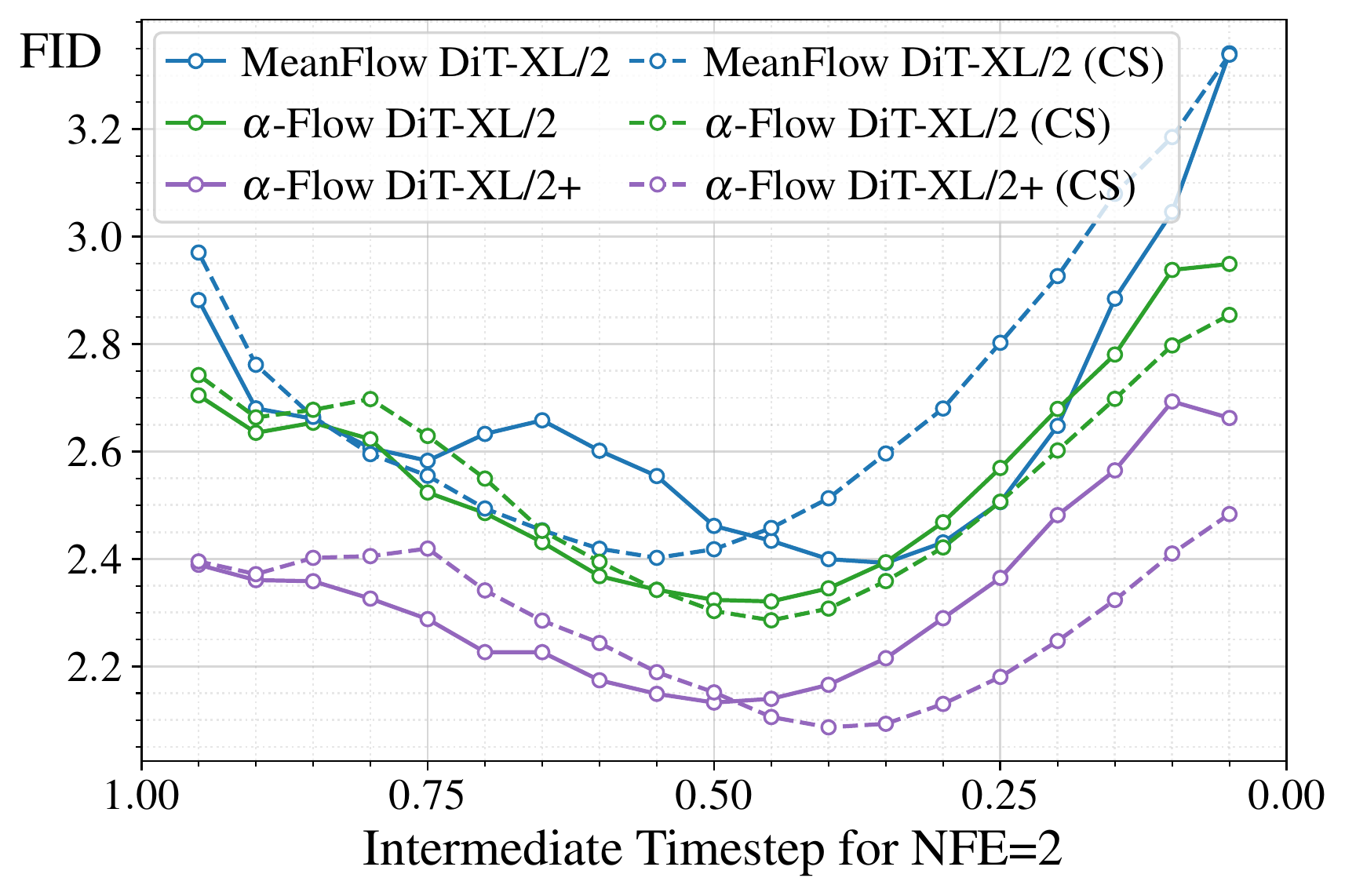}
  \caption{Comparing ODE vs consistency sampling for MeanFlow and \modelname models.}
  \label{fig:sampling}
\end{minipage}%
\hfill
\begin{minipage}{.48\textwidth}
\centering
\includegraphics[height=4.2cm]{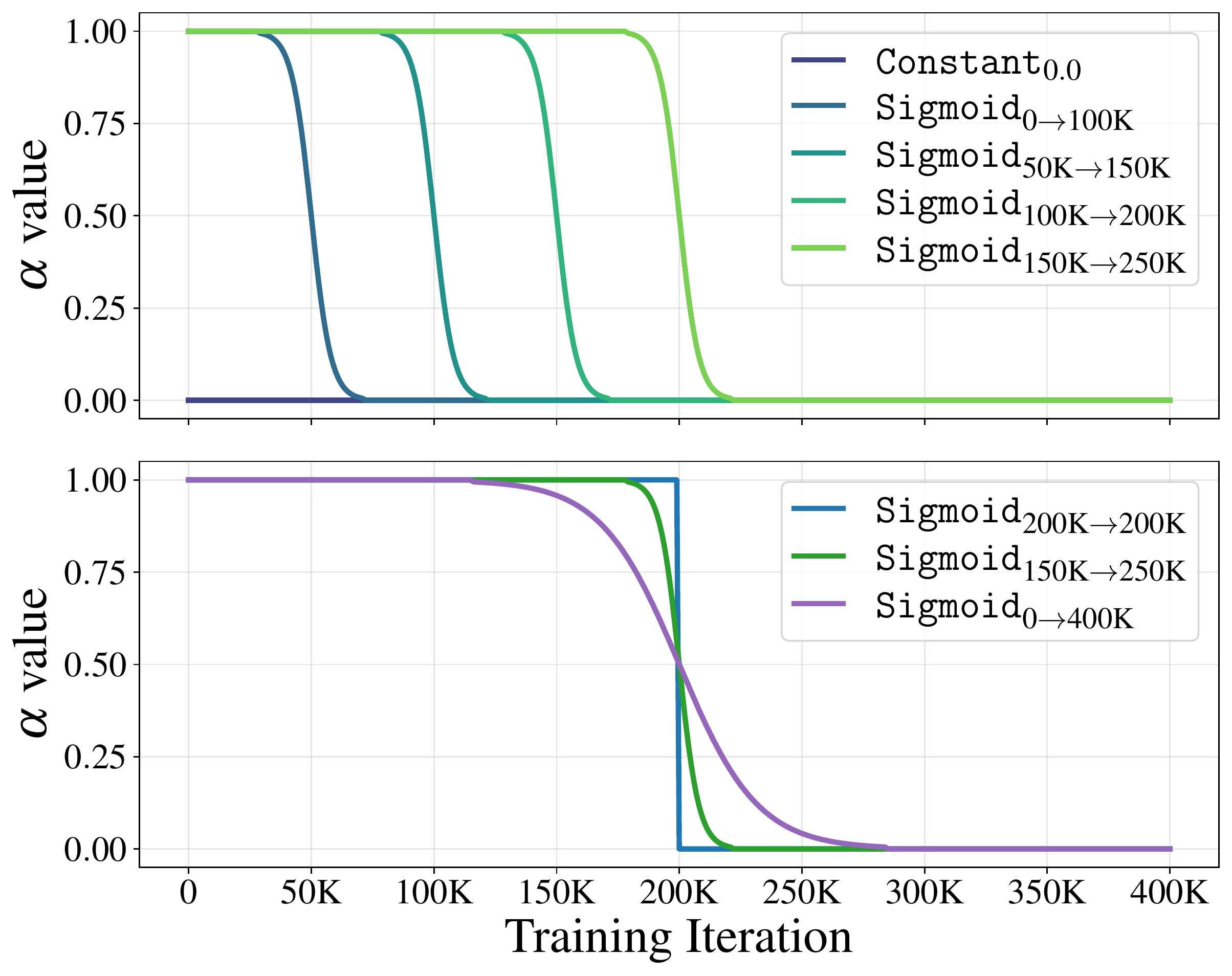}
\caption{Visualization of \rationame schedule.}
\label{fig:schedule_vis}
\end{minipage}
\vspace{-0.1in}
\end{figure}

\vspace{-0.2in}
In \Cref{tab:ablation_alphaflow} (a), we evaluate our \modelname framework trained with various sigmoid schedules, as visualized in \Cref{fig:schedule_vis}. 
For these experiments, the \bFMname ratio is fixed at 25\%.
We first analyze the impact of the \uFMname pretraining duration. 
By fixing $k_e - k_s$ to 100K iterations, we progressively increase $k_s$ from 0K to 150K. 
As the pretraining duration increases, \modelname's performance consistently improves across all metrics. 
The best-performing schedule, $\SigmoidSch{150}{250}$, significantly outperforms the baseline MeanFlow ($\mathtt{Constant}_{0.0}$).
This suggests that \textit{optimizing \uFMname is more crucial than optimizing MeanFlow in the early training stages for achieving superior few-step flow modeling}.
This finding aligns with our empirical analysis, which shows that because the gradients of the \uFMname and consistency losses conflict, it is more efficient to exclusively optimize the \uFMname objective for faster initial convergence.

Next, we investigate the effect of the transition duration. 
With the midpoint $(k_s + k_e)/2$ fixed at 200K iterations, we vary the total transition iterations from 0 to 400K. 
Our results indicate that a longer, smoother transition leads to better generation quality. 
This highlights the importance of gradually reducing the bias of the training objective by smoothly transitioning between \uFMname and MeanFlow.

\vspace{-0.2in}
\inlinesection{Flow matching ratio}

In \Cref{tab:ablation_alphaflow} (b), we compare our \modelname framework with the MeanFlow baseline across various \BFMname ratios ($\% r = t$). 
Our results show that \modelname consistently outperforms MeanFlow for all evaluated ratios, confirming the effectiveness of our proposed method.
A key finding is that \modelname achieves its best 1-NFE performance at a relatively low \bFMname ratio. 
Specifically, it reaches the best FID of 40.0 at 25 \% of $ r = t$ and the best FDD of 781.0 at 50 \% of $ r = t$, while MeanFlow requires a higher ratio of 75\% to achieve its best FID of 43.1 and FDD of 819.2. This aligns with our motivation: by pretraining on \uFMname, \modelname is less reliant on the \BFMname objective and can focus more on the overall MeanFlow objective, leading to superior one-step generation quality.

Furthermore, we observe that for \modelname, the \bFMname ratio presents a clear trade-off between 1-NFE and 2-NFE performance.
For instance, the 75\% ratio yields worse NFE=1 but better NFE=2 generation results compared to the 50\%-ratio version.
This indicates that a higher proportion of flow matching improves the model's ability to generate images in a slightly higher number of steps. 
\vspace{-0.1in}
\inlinesection{Sampling} As shown in \Cref{fig:sampling}, we compare ODE sampling (solid line) and consistency sampling (dotted line) for 2-NFE generation across different intermediate sampling timesteps, using MeanFlow-XL/2, \modelname-XL/2, and \modelname-XL/2+. The results show that consistency sampling yields better generation performance for both \modelname-XL/2 and \modelname-XL/2+, achieving the best FID scores of 2.09 at timestep 0.4 and 2.28 at timestep 0.45, respectively. In contrast, ODE sampling performs better for MeanFlow-XL/2, which attains its best FID of 2.39 at timestep 0.35. In \Cref{tab:imagenet256}, we select intermediate sampling timesteps that balance FID and FDD; see \Cref{tab:imagenet-configs-dit} for details.


\begin{figure}[t] 

\end{figure}

\section{Conclusion}
\label{sec:conclusion}

Our work provided a principled analysis of the MeanFlow framework, analyzing its objective and establishing the necessity of flow matching supervision during training.
Motivated by this understanding, we proposed the $\alpha$-Flow objective as a generalization of MeanFlow loss, allowing us to train consistently stronger few-step image generation models from scratch.



\section{Reproducibility Statement}
\label{sec:reproducibility}

We are committed to ensuring the reproducibility of our results. 
To this end, we include all the necessary implementation details in \Cref{sec:impl-details}, ensuring that our methodology can be faithfully reproduced. 
We will publicly release our source training, inference, and evaluation code, as well as the pre-trained checkpoints for \INFull. 

\bibliography{references}
\bibliographystyle{iclr2026_conference}

\clearpage
\appendix
\section{Related Work}
\label{sec:related-work}




\inlinesection{Diffusion Models} 
Diffusion models have become a dominant paradigm in generative modeling for vision domains~\citep{NonEqThermodynamics,NCSN,DDPM,DDIM,SBGM,ADM}. 
The classical diffusion framework defines a forward noising process and a corresponding reverse process that a model learns to approximate. 
Early works such as DDPM~\citep{DDPM} and score-based generative modeling~\citep{NCSN} demonstrated high-quality image generation, later extended to continuous-time SDEs and ODEs~\citep{SBGM}. 
\citep{ADM} further improved sample fidelity with larger architectures and classifier guidance. 
More recently, the community has explored \emph{flow-based} parameterizations that directly learn continuous velocity fields~\citep{FlowStraightAndFast,flowmatching,StochasticInterpolants}. 
These flow matching approaches simplify training, unify score- and likelihood-based models, and are used in large-scale systems such as Stable Diffusion 3~\citep{SD3}.

\inlinesection{Few-step Diffusion} 
Despite their quality, diffusion models are computationally expensive due to iterative sampling. 
A large body of work accelerates sampling to a few steps or even one step. 
Distillation-based approaches include progressive distillation~\citep{ProgressiveDistillation, TRACT}, and often incorporate adversarial objectives~\citep{DMD, DMD2, SID, ADD}.
UCGM~\citep{UCGM} develops a unified training scheme for multi-step and few-step diffusion-based methods.

A closer research direction (which our method follows as well) includes the methods which are trained from scratch and support few- and even one-step generation by design.
Consistency Models (CMs)~\citep{CMs} learn to map noisy inputs directly to clean data by enforcing self-consistency. 
Extensions improve stability and scalability~\citep{iCT, sCM, EasyCM}. 
\emph{Trajectory-based} methods learn the dynamics of the entire denoising process, enabling arbitrary jumps along the diffusion path. 
PCM~\citep{PCM} scale consistency distillation to large scale models and optimize with preselected time intervals.
Shortcut diffusion models~\citep{ShortcutModels} learn direct mappings with shortcut constraints. 
MeanFlow~\citep{MeanFlow} predicts time-averaged velocities with continuous consistency, while \cite{SplitMF} explore this idea for discrete consistency. 
Hybrid approaches combine consistency and flow matching: Consistency-FM~\citep{ConsistencyFM} enforces velocity self-consistency, FACM~\citep{FACM} anchors consistency to flow objectives, and IMM~\citep{IMM} matches the output distributions via moment matching instead of exact outputs. 
Consistency Trajectory Models (CTM)~\citep{CTM} generalize consistency training to support transitions between any two timesteps, combining one-step generation with progressive refinement. 
Transition Models (TiM)~\citep{TransitionModel} derive an exact continuous-time dynamics equation for arbitrary-step transitions. 
These methods achieve one- to few-step sampling with steadily improving fidelity.

\section{Limitations}
\label{sec:limitations}

\begin{itemize}[leftmargin=*]
\item Our $\alpha$-Flow loss enables high-quality training of discrete MeanFlow models without requiring JVP computation. However, in practice, the continuous objective (i.e., setting $\alpha \to 0$) remains important, likely due to the bias–variance trade-off inherent in the consistency objective~\citep{CMs,iCT}.
\item We occasionally observed unstable training in large-scale models with guidance integration, both for the vanilla MeanFlow model and our $\alpha$-Flow variant. Thus, our framework should not be viewed as a silver bullet for addressing the well-known instability issues of consistency models~\cite{EasyCM}.
\item The $\alpha$-Flow objective uses pure flow matching supervision up to $k_s$ iterations, after which the consistency objective is applied. Before this point, the model’s few-step performance is weak, which can make progress harder to monitor.
\item Our gradient analysis provides actionable insights but remains empirical; it does not fully explain, from a theoretical perspective, why flow matching is so critical for consistency.
\item Although we motivate larger batch sizes for fine-tuning by the high variance of the consistency loss, the observed improvements (see \Cref{tab:ablation-batch-size}) may instead reflect that small batches are more sensitive to hyperparameters~\citep{SmallBatchLLM}, and that beyond a certain size, batch-size scaling exhibits diminishing returns~\citep{CriticalBatchSize}.
\end{itemize}

\section{Failed Experiments}
\label{sec:failed-experiments}

We also wish to share with the community several experiments that did not succeed during the course of this project.  
Some of these directions were likely underexplored on our side, while others may represent genuine dead-ends.  
Nevertheless, we believe documenting them may serve as a useful reference for future work.  

\begin{itemize}[leftmargin=*]
\item We devoted several weeks to exploring decomposed training of the MeanFlow objective with individually tuned weighting functions for each term, drawing inspiration from EDM~\cite{EDM} to map out the design space. Unfortunately, every configuration we attempted produced worse results than the default adaptive loss heuristic, which was a particularly frustrating outcome.  
\item Consistency sampling (see \Cref{fig:sampling}) did not provide the improvements we had anticipated. Interestingly, the optimal midpoint consistently emerged at $\approx 0.5$, which coincides with the default MeanFlow setting. We suspect this effect is related to the training distribution, which has a mode slightly lower 0.5. Following the original work, we employed a logit-normal distribution with location parameter $-0.4$. 
\item We experimented with LoRA fine-tuning and introduced separate prediction heads for vanilla velocity and mean velocity. Neither approach yielded promising results.  
\item We conducted roughly 50 ablations on the train-time noise schedule for vanilla MeanFlow models. None resulted in noticeably better performance, even when factorizing the joint distribution $p(t, r)$ into $p(t)p(r|t)$ and exploring alternative supervision distributions for flow matching in parallel.  
\item We investigated additional representation alignment losses~\cite{REPA} with the aim of accelerating convergence in MeanFlow models. However, the observed gains were insufficient to justify the added complexity of the training framework.  
\item We also experimented with different EMA schedules, but these attempts did not lead to meaningful improvements.  
\end{itemize}

\section{Proofs of things}
\subsection{Loss decomposition}
\label{app:proof_decomp}


\begin{proof}

The MeanFlow loss is given by:
\begin{align}
\LMFc(\bm \theta) &= \mathbb{E}_{t,r, \vz_t}\left[\left\Vert\vu_{\bm \theta}(\vz_t, r, t) - \vv_t + (t - r) \frac{\text{d} \vu_{\bm \theta^{-}}(\vz_t, r, t)}{\text{d} t} \right\Vert^2_2\right] \\
\intertext{(unpacking the norm and regrouping terms yields)}
&= \underbrace{\mathbb{E}_{t,r, \vz_t}\left[\left\Vert\vu_{\bm \theta}(\vz_t, r, t) - \vv_t \right\Vert^2_2\right]}_{\LuFM\left(\bm \theta\right)} +\underbrace{\mathbb{E}_{t,r, \vz_t}\left[2\cdot \left(t - r\right) \cdot \vu_{\bm \theta}^{\top}(\vz_t, r, t) \frac{\text{d} \vu_{\bm \theta^{-}}(\vz_t, r, t)}{\text{d} t} \right]}_{  \LCFc\left(\bm \theta\right)} \\
&+\underbrace{\mathbb{E}_{t,r, \vz_t}\left[-2 \left(t - r\right) \cdot \vv^{\top}(\vz_t, t|\vx) \frac{\text{d} \vu_{\bm \theta^{-}}(\vz_t, r, t)}{\text{d} t}  + \left(t - r\right)^2\left\Vert\frac{\text{d} \vu_{\bm \theta^{-}}(\vz_t, r, t)}{\text{d} t} \right\Vert^2_2\right]}_{\text{Does not depend on $\param$}}
\end{align}
\end{proof}

\subsection{\texorpdfstring{$\LMFd$}{alpha-Flow} loss unification}
\label{app:proof_unify}

\begin{proof}[Proof of \cref{thm:unified-loss}]
The proof for flow matching and shortcut models is straightforward.
We will only show the proof for the third bullet point.
For brevity, let's set $\Delta t = t - s$ and $\stepratio = \dfrac{\Delta t}{t - r}$.

\begin{equation}
\label{eq:loss-val-MFd-MFc}
\begin{aligned}
\LMFd(\bm \theta) &=\mathbb{E}_{t,r, \vz_t }\left[\omegaMFd \cdot\left\Vert\vu_{\bm \theta}(\vz_t, r, t) - \frac{\Delta t}{t -r} \cdot \vv_t - \right.\right.\\
&\qquad\qquad\qquad\qquad\qquad\qquad\left.\left.
\frac{t - \Delta t -r}{t -r}\vu_{\bm \theta^{-}}\left(\vz_{t-\Delta t}, r, t-\Delta t\right)\right\Vert^2_2\right], \\
&\stackrel{(i)}{=}\mathbb{E}_{t,r, \vz_t}
\left[\omegaMFd \cdot\left\Vert
\vu_{\bm \theta}(\vz_t, r, t) - \frac{\Delta t}{t -r} \cdot \vv_t  - \frac{t - \Delta t -r}{t -r} \cdot \right.\right.\\
&\qquad\qquad\qquad\left.\left. \left(\vu_{\bm \theta^{-}}\left(\vz_t, r, t\right) - \frac{\text{d} \vu_{\bm \theta^{-}}(\vz_t, r, t)}{\text{d} t}\Delta t + \mathcal{O}\left(\Delta^2 t\right) \right)
\right\Vert^2_2\right], \\
&\stackrel{(ii)}{=}\mathbb{E}_{t,r, \vz_t}
\left[\omegaMFd \cdot\left\Vert
\vu_{\bm \theta}(\vz_t, r, t) - \vu_{\bm \theta^{-}}(\vz_t, r, t) -\frac{\Delta t}{t -r} \cdot\right.\right.\\
&\qquad\qquad\qquad\left.\left. \left(\vv_t - (t - r)\frac{\text{d} \vu_{\bm \theta^{-}}(\vz_t, r, t)}{\text{d} t} - \vu_{\bm \theta^{-}}(\vz_t, r, t) +  \mathcal{O}\left(\Delta^2 t\right) \right)
\right\Vert^2_2\right], \\
\end{aligned}
\end{equation}
where $(i)$ uses the Taylor expansion over $\vu_{\bm \theta^{-}}\left(\vz_{t-\Delta t}, r, t-\Delta t\right)$: $$\vu_{\bm \theta^{-}}\left(\vz_{t-\Delta t}, r, t-\Delta t\right) = \vu_{\bm \theta^{-}}\left(\vz_t, r, t\right) - \dfrac{\text{d} \vu_{\bm \theta^{-}}(\vz_t, r, t)}{\text{d} t}\Delta t + \mathcal{O}\left(\Delta^2 t\right),$$
and $(ii)$ uses the fact that $\dfrac{\text{d} \vu_{\bm \theta^{-}}(\vz_t, r, t)}{\text{d} t} \Delta ^2t = \mathcal{O}\left(\Delta^2 t\right)$. Thus, 
\begin{equation}
    \begin{aligned}
        &\lim_{\stepratio \rightarrow 0 }\nabla_{\bm \theta}\LMFd(\bm \theta)
        = \lim_{\Delta t \rightarrow 0 }\nabla_{\bm \theta}\LMFd(\bm \theta)
        \\= &\lim_{\Delta t \rightarrow 0 }\mathbb{E}_{t,r, \vz_t}
        \left[2 \cdot \omegaMFd \cdot\nabla^{\top}_{\bm \theta}\vu_{\bm \theta}(\vz_t, r, t) \cdot \left(
        \vu_{\bm \theta}(\vz_t, r, t) - \vu_{\bm \theta^{-}}(\vz_t, r, t) -\frac{\Delta t}{t -r} \cdot\right.\right.\\
        &\qquad\qquad\qquad\left.\left. \left(\vv_t - (t - r)\frac{\text{d} \vu_{\bm \theta^{-}}(\vz_t, r, t)}{\text{d} t} - \vu_{\bm \theta^{-}}(\vz_t, r, t) +  \mathcal{O}\left(\Delta^2 t\right) \right)
        \right)\right], \\
        = &\lim_{\Delta t \rightarrow 0 }\mathbb{E}_{t,r, \vz_t}
        \left[2 \cdot \omegaMFd \cdot\nabla^{\top}_{\bm \theta}\vu_{\bm \theta}(\vz_t, r, t) \cdot \left(-\frac{\Delta t}{t -r} \right)\cdot\right.\\
        &\qquad\qquad\qquad\left. \left(\vv_t - (t - r)\frac{\text{d} \vu_{\bm \theta^{-}}(\vz_t, r, t)}{\text{d} t} - \vu_{\bm \theta^{-}}(\vz_t, r, t) +  \mathcal{O}\left(\Delta^2 t\right)\right)\right], \\
        = &\mathbb{E}_{t,r, \vz_t}
        \left[2 \cdot \left(\omegaMFd \cdot \frac{\Delta t}{t -r}\right) \cdot \nabla^{\top}_{\bm \theta}\vu_{\bm \theta}(\vz_t, r, t) \cdot \left( \vu_{\bm \theta^{-}}(\vz_t, r, t) - \vv_t + (t - r)\frac{\text{d} \vu_{\bm \theta^{-}}(\vz_t, r, t)}{\text{d} t}\right)\right], \\
        = &\nabla_{\bm \theta}\LMFc(\bm \theta),\\
    \end{aligned}
\end{equation}
\end{proof}

\begin{proof}[Proof of equivalence with consistency model]
    By setting $\vvoffset = \vv_t, r = 0$ and $\vu_{\bm \theta}(\vz_t, 0, t) = \left(\vz_t - \vf_{\bm \theta}(\vz_t, t)\right)/t$, $\Delta t = t - s$ and $\stepratio = \dfrac{\Delta t}{t}$, we have:
    \begin{equation}
        \begin{aligned}
            \LMFd(\bm \theta) &=\mathbb{E}_{t,r, \vz_t}\left[\dfrac{t}{\Delta t}\cdot\left\Vert\vu_{\bm \theta}(\vz_t, r, t) - \frac{\Delta t}{t -r} \cdot \vv_t - \right.\right.\\
            &\qquad\qquad\qquad\qquad\qquad\qquad\left.\left.
            \frac{t - \Delta t -r}{t -r}\vu_{\bm \theta^{-}}\left(\vz_{t-\Delta t}, r, t-\Delta t\right)\right\Vert^2_2\right], \\
            &\stackrel{(i)}{\Rightarrow}\mathbb{E}_{t,\vz_t}\left[\dfrac{t}{\Delta t} \cdot\left\Vert\frac{\vz_t - \vf_{\bm \theta}(\vz_t, t)}{t} - \frac{\Delta t}{t} \cdot \vv_t - \right.\right.\\
            &\qquad\qquad\qquad\qquad\qquad\qquad\left.\left.
            \frac{t - \Delta t}{t}\frac{\vz_{t-\Delta t} - \vf_{\bm \theta^{-}}(\vz_{t-\Delta t}, t-\Delta t)}{t-\Delta t}\right\Vert^2_2\right], \\
            &\stackrel{(ii)}{=} \mathbb{E}_{t,\vz_t}\left[\dfrac{1}{t\Delta t} \cdot \left\Vert\vf_{\bm \theta}(\vz_t, t) - \vf_{\bm \theta^{-}}(\vz_{t-\Delta t}, t-\Delta t)\right\Vert^2_2\right] \stackrel{(iii)}{=}  \LCTd(\bm \theta), \\
        \end{aligned}
    \end{equation}
    where $(i)$ plug in the reparameterization and $r=0$, $(ii)$ uses the fact that $\vz_t = \vz_{t - \Delta t} + \Delta t \cdot \vv_t$. Thus $\LMFd(\bm \theta)$ could be reparameterized to $\LCTd(\bm \theta)$ with a loss weighting function $\dfrac{1}{t\Delta t}$. Since the discrete CT uses timestep partition to determine $t$ and $\Delta t$, the $(iii)$ holds for a special timestep partition when $\Delta t = \alpha \cdot t$ given a fixed $\alpha$.  From Theorem 2 in \cite{flowmatching}, because $\vu_{\bm \theta^{-}}(\vz_t, r, t)$ is independent of $\bm \theta$, we have:
    \begin{equation}
    \begin{aligned}
        \LMFc(\bm \theta)
        &= \mathbb{E}_{t,r, \vz_t}\left[\left\Vert\vu_{\bm \theta}(\vz_t, r, t) - \vv_t + (t - r) \frac{\text{d} \vu_{\bm \theta^{-}}(\vz_t, r, t)}{\text{d} t} \right\Vert^2_2\right], \\
        &= \mathbb{E}_{t,r, \vz_t}\left[\left\Vert\vu_{\bm \theta}(\vz_t, r, t) - \vv(\vz_t, t) + (t - r) \frac{\text{d} \vu_{\bm \theta^{-}}(\vz_t, r, t)}{\text{d} t} \right\Vert^2_2\right] + C, \\
    \end{aligned}
    \end{equation}
    with $C$ a constant independent of $\bm \theta$. Thus,
    \begin{equation}
        \begin{aligned}
            &\nabla_{\bm \theta}\LMFc(\bm \theta) \\
            &= \mathbb{E}_{t,r, \vz_t}
        \left[2 \cdot \nabla^{\top}_{\bm \theta}\vu_{\bm \theta}(\vz_t, r, t) \cdot \left( \vu_{\bm \theta}(\vz_t, r, t) - \vv(\vz_t, t) + (t - r)\frac{\text{d} \vu_{\bm \theta^{-}}(\vz_t, r, t)}{\text{d} t}\right)\right], \\
        &\stackrel{(i)}{\Rightarrow} \mathbb{E}_{t,\vz_t}
        \left[2 \cdot \nabla^{\top}_{\bm \theta} \frac{-\bm f_{\bm \theta}(\vz_t, t)}{t} \cdot \left( \frac{\vz_t - f_{\bm \theta}(\vz_t, t)}{t} - \vv(\vz_t, t) + \right.\right.\\
        &\qquad\qquad\qquad\left.\left. \left(\vv(\vz_t, t) - \frac{\text{d} \vf_{\bm \theta^{-}}\left(\vz_{t},  t\right)}{\text{d} t}\right) - \frac{\left(\vz_t - f_{\bm \theta^{-}}(\vz_t, t)\right)}{t}\right)\right], \\
        &= \mathbb{E}_{t,\vz_t}\left[2 \cdot \frac{1}{t} \cdot \nabla^{\top}_{\bm \theta} \bm f_{\bm \theta}(\vz_t, t)\frac{\text{d} \vf_{\bm \theta^{-}}\left(\vz_{t},  t\right)}{\text{d} t}\right] = \nabla_{\bm \theta}\LCTc(\bm \theta),
        \end{aligned}
    \end{equation}
    where $(i)$ plug in the reparameterization and $r=0$, and use the fact that:
    \begin{equation}
        \frac{\text{d} \vu_{\bm \theta^{-}}(\vz_t, r=0, t)}{\text{d} t} = \frac{1}{t^2} \left(t \left(\vv(\vz_t, t) - \frac{\text{d} \vf_{\bm \theta^{-}}\left(\vz_{t},  t\right)}{\text{d} t}\right) - \left(\vz_t - \bm f_{\bm \theta^{-}}(\vz_t, t)\right)\right)
    \end{equation}
    Thus $\nabla_{\bm \theta}\LMFc(\bm \theta)$ could be reparameterized to $\LCTc(\bm \theta)$ with a loss weighting function $\dfrac{1}{t}$.
\end{proof}
\section{Analysis details}\label{app:analysis-details}

The detailed implementation of DiT-B/2 is provided in \Cref{tab:imagenet-configs-dit}, where we adopt the DiT-B/2-non-cfg setting. For loss evaluation, at each checkpoint we use a batch size of 128 and run 1000 iterations to compute the mean loss along with its 5\% and 95\% percentiles, which are reported in the figure. To measure the cosine similarity between different losses, we calculate $\nabla\LuFM$, $\nabla\LbFM$, and $\nabla\LCFc$ on the same batch and then compute their pairwise cosine similarities. This procedure is also repeated over 1000 iterations to obtain the mean similarity and its 5\% and 95\% percentiles, as shown in the figure.

\section{Implementation details}
\label{sec:impl-details}

Implementation details are shown in \Cref{tab:imagenet-configs-dit}.


\begin{table}[h]
\small
\centering
\caption{Configurations on ImageNet 256$\times$256. B/2-non-cfg is our ablation and analysis model in the main text.
}
\tablestyle{4pt}{1}

\begin{tabular}{lcccc}
\toprule
Configs & DiT-B/2-non-cfg
& DiT-B/2 & DiT-XL/2
& DiT-XL/2+ \\
\midrule
\textit{Network Architectures} \\
Params (M) & $131$ & $131$ & $676$ & $676$ \\
FLOPs (G)  & $23.1$ & $23.1$ & $119.0$ & $119.0$ \\
Depth      & $12$  & $12$ & $28$ & $28$\\
Hidden dim & $768$ & $768$ & $1152$ & $1152$\\
Heads      & $12$  & $12$ & $16$ & $16$\\
Patch size  & $2{\times}2$ & $2 \times 2$ &  $2 \times 2$ &  $2 \times 2$\\

\midrule
\textit{Training hyperparameters} \\
Training steps & 400K & 1.2M & 1.2M & 1.2M\\
Batch size for training & 256 & 256 & 256 & 256 \\
Fine-tuning steps & \emptytable & \emptytable & \emptytable & 75K\\
Batch size for fine-tuning & \emptytable & \emptytable & \emptytable & 1024\\
Dropout
            & \multicolumn{4}{c}{$0.0$} \\
Optimizer
            & \multicolumn{4}{c}{Adam \cite{adam}} \\
lr schedule
            & \multicolumn{4}{c}{constant} \\
lr & \multicolumn{4}{c}{$0.0001$} \\
Adam $(\beta_1, \beta_2)$
            & \multicolumn{4}{c}{$(0.9, 0.95)$} \\
Weight decay & \multicolumn{4}{c}{0.0} \\
EMA half-life & \multicolumn{4}{c}{6931} \\
Gradient clipping norm & \multicolumn{4}{c}{16} \\
Autoencoder used & \multicolumn{4}{c}{sd-vae-ft-ema} \\
\midrule
\textit{\modelname  hyperparameters} \\
Ratio of $r = t$ & \Cref{tab:ablation_alphaflow} (b)  & $25\%$ & $50\%$ & $50\%$ \\
$(r, t)$ sampler & \multicolumn{4}{c}{logitnorm(--0.4, 1.0)} \\
$\vvoffset$ & \Cref{tab:discrete_ablation} (a) & $\vv_t$ & $\vv_t$ & $\vv_t$ \\
Whether to use EMA for $\vu_{\bm \theta^{-}}$ & \Cref{tab:discrete_ablation} (a) & No & No & No \\
Adaptive weight & \Cref{tab:discrete_ablation} (b) & \multicolumn{3}{c}{$\omega = \alpha/\left(||\Delta||_2^2 + c\right)$} \\
\midrule
\textit{Schedule of $\stepratio$} \\
$\gamma$  & 25 & 25 & 25 \\
$k_s$ & \multirow{2}{*}{\Cref{tab:ablation_alphaflow} (b)} & 0 & 600K & 600K \\
$k_e$ & \ & 1.2M & 1M & 1M \\
 $\eta$ & \Cref{tab:discrete_ablation} (c) & $5 \times 10^{-3}$ & $5 \times 10^{-3}$ & $5 \times 10^{-3}$ \\
 \midrule
\textit{CFG training} \\
$\omegacfg$ & \emptytable & 1.0 & 0.2 & 0.2 \\
$\kappacfg$ & \emptytable & 1.0 & 0.92 & 0.92 \\
CFG triggered if $t$ is in & \emptytable & $[0.0, 1.0]$ & $[0.0, 0.75]$ & $[0.0, 0.75]$ \\
Whether use EMA for CFG & \emptytable & No & No & No \\
\midrule
\textit{2-NFE Sampling} \\
Method & ODE  & ODE & consistency   & consistency  \\ 
Intermediate timestep  & 0.5 & 0.5 & 0.55 & 0.5 \\
\bottomrule

\end{tabular}
\label{tab:imagenet-configs-dit}
\vspace{1em}
\end{table}

\section{Additional Experiments}


\begin{table}[t] 
\centering
\resizebox{.47\linewidth}{!}{
\begin{minipage}{.6\linewidth}
\begin{algorithm}[H]
\caption{$\alpha$-Flow: Sampling}
\label{alg:code_sample}
\begin{lstlisting}[mathescape]
# 1 = t1 > t2 > ... > tN = 0 :sequence of timesteps
z = randn_like(x)
for n in range(N):
    m = n + 1
    if consistency_sampling:
        z = z - tn * fn(z, r=0, t=tn)
        z = z + tm * randn_like(x)
    elif ODE_sampling:
        z = z - (tn - tm) * fn(z, r=tm, t=tn)
    
\end{lstlisting}
\end{algorithm}

    

\end{minipage}
}
\hfill
\begin{minipage}{.5\textwidth}
\centering
\tablestyle{2pt}{1.0}
\small
\begin{tabular}{l cccc}
    \toprule
    \multirow{2}{*}{Batch Size} & \multicolumn{2}{c}{NFE 1} & \multicolumn{2}{c}{NFE 2} \\
     \cmidrule(lr){2-3} \cmidrule(lr){4-5}
     & \FID & \FDD & \FID & \FDD \\
    \midrule
    256 & 3.13 & 167.2 & 2.31 & 97.1\\
    512 & \textbf{3.05} & \textbf{164.3} & 2.21 & 95.2\\
    1024& 3.06 & 165.7 & 2.16 & \textbf{93.4}\\   
    2048& 3.29 & 169.6 & \textbf{2.10} & 96.6\\   
    4096& 3.13 & 168.9 & 2.16 & 95.1\\   
    \bottomrule
\end{tabular}
\caption{Ablation study over the fine-tuning batch size using the data distribution over class labels.}
\label{tab:ablation-batch-size}
\end{minipage}

\end{table}

\subsection{Ablation study over batch size}

Training diffusion/flow-based models can be challenging due to the high variance of their gradients. Past research \cite{IMM, EDM, EDM2} often used large batch sizes (1024 or even 4096) to mitigate this issue. In this section, we fine-tune a MeanFlow-XL/2 model (with implementation details in \Cref{tab:imagenet-configs-dit}) for an additional 60 epochs using a large batch size.

As shown in \Cref{tab:ablation-batch-size}, a batch size of 512 achieved the best 1-NFE FID of 3.05 and FDD of 164.3. A batch size of 1024, however, yielded the best FDD of 93.4. Overall, a batch size of 1024 performed well across all metrics, so we designate this configuration as MeanFlow-XL/2+. The same setting is applied to fine-tune the MeanFlow-XL/2 model, leading to the MeanFlow-XL/2+ results in \Cref{tab:imagenet256}. Our proposed \modelnamexlp model outperforms MeanFlow-XL/2+ in several key metrics: 1-NFE FID (2.58 vs. 3.06), 1-NFE FDD (148.4 vs. 165.7) and 2-NFE FID (2.15 vs. 2.16), only worse in 2-NFE FDD (96.8 vs. 93.4).
These results demonstrate the overall effectiveness of our \modelname method.
Notably, the results in \Cref{tab:ablation-batch-size} are obtained using labels sampled from the ImageNet dataset distribution, whereas the results in \Cref{tab:imagenet256} use randomly generated labels.
In general, sampling labels from the ImageNet distribution leads to lower FID scores compared to using random labels.


\subsection{Ablation study over \texorpdfstring{$\alpha$-Flow}{alphaFlow} design space}
\label{app:ablation-discrete-meanflow}

\begin{table*}[ht]
\centering
\begin{minipage}{0.48\textwidth}
    \centering
    \tablestyle{2pt}{1.}
    \small
    \begin{tabular}{x{40}x{48}x{25}x{40}}
        $\vvoffset$ & $\vu_{\bm \theta^{-}}$ &\FID & \FDD \\
        \midrule
        $\vu_{\bm \theta^{-}}$ & EMA & 188.1 & 1761.6\\
        $\vu_{\bm \theta^{-}}$ & Non-EMA & 319.0 & 4009.9\\
        $\vv_t$ & EMA & 202.8 & 1832.3\\
        \rowcolor[gray]{0.8}
        $\vv_t$ & Non-EMA & 59.2 & 964.6\\
    \end{tabular}
    \caption*{(a) \textbf{Reformulate the training objective.}}
\end{minipage}
\hfill
\begin{minipage}{0.48\textwidth}
    \centering
    \tablestyle{2pt}{1.}
    \small
    \begin{tabular}{L{102}x{25}x{40}}
        Loss weight & \FID & \FDD \\
        \midrule
        $\omega = 1$ & 59.2 & 964.6 \\
        $\omega = 1/\left(||\Delta||_2^2 + c\right)^{0.5}$ & 55.0 & 918.5 \\
        $\omega = 1/\left(||\Delta||_2^2 + c\right)$ & 52.2 & 883.6 \\
        \rowcolor[gray]{0.8}
        $\omega = \alpha/\left(||\Delta||_2^2 + c\right)$ & \textbf{49.7} & \textbf{845.2} \\
    \end{tabular}
    \caption*{(b) \textbf{Adaptive loss.}}
\end{minipage}
\par\bigskip
\begin{minipage}{0.48\textwidth}
    \centering
    \tablestyle{2pt}{1.}
    \small
    \begin{tabular}{x{96}x{25}x{40}}
        $\stepratio$ & \FID & \FDD \\
        \midrule
        $10^{-2}$ & 49.7 & 845.2\\
        \rowcolor[gray]{0.8}
        $5 \times 10^{-3}$ & \textbf{46.2} & 860.8 \\
        $2 \times 10^{-3}$ & 50.3 & \textbf{833.0} \\
        $1 \times 10^{-3}$ & 57.2 & 863.7 \\
    \end{tabular}
    \caption*{\textbf{(c) \Rationame.}}
\end{minipage}
\hfill
\begin{minipage}{0.48\textwidth}
    \centering
    \tablestyle{2pt}{1.}
    \small
    \begin{tabular}{L{96}x{25}x{40}}
          Method & \FID & \FDD \\
          \hline
          Shortcut Model & 59.8 & 1017.3 \\ 
          $\Tilde{\vv} = \vv(\vz_t, t|\vx)$ & 59.2 & 964.6 \\
          + Adaptive loss & 49.7 & 845.2 \\
          + $\alpha = 0.005$  & 45.6 & 857.8 \\
          \hline
          MeanFlow & 43.3 & 822.3 \\
    \end{tabular}
    \caption*{\textbf{(d) Overall ablation study.}}
\end{minipage}
\caption{Ablation study over \modelname.}
\label{tab:discrete_ablation}
\end{table*}


This section contains an ablation study on \modelname, specifically for $\alpha \in (0, 1)$. 
We use a DiT-B/2-non-cfg model (see Table \ref{tab:imagenet-configs-dit}) that is pre-trained on flow matching for 200k iterations and then fine-tuned on \modelname for another 200k iterations. 
Across all experiments, $\alpha$ remains a constant, and the ratio of $r = t$ is 25 \%.

\inlinesection{Training objective} Here, we set $\alpha = 10^-2$. Table \ref{tab:discrete_ablation}(a) shows that the model only converge when $\vvoffset$ was set to $\vv_t$ and without using EMA for $\vu_{\bm \theta^{-}}$. 
This is a key difference from Shortcut Models \cite{ShortcutModels}, which set $\vvoffset = \vu_{\bm \theta^{-}}$. We suspect their objective only works when $\stepratio$ is larger (e.g., 0.5).

\inlinesection{Adaptive loss} \cite{MeanFlow} uses an adaptive weight: $\omega = 1/\left(||\Delta||_2^2 + c\right) = 1/\left(\LMFc + c\right)$. From \Cref{eq:loss-val-MFd-MFc}, we could derive $\lim_{\alpha \rightarrow 0}\LMFd = \alpha \LMFc$. 
When $\alpha$ is close 0, we approximate $\LMFc$ as $\LMFd/\alpha$. 
This gives us a new adaptive weight, $\omega  = 1/\left(\LMFd/\alpha + c\right) \approx \alpha/\left(\LMFd + c\right) = \alpha/\left(||\Delta||_2^2 + c\right)$ as both $c$ and $\alpha$ is very small. 
As shown in \Cref{tab:discrete_ablation}(b), this new weight performs better empirically, especially compared to the original MeanFlow adaptive weight.

\inlinesection{\Rationame} Ablating the  $\stepratio$ in \Cref{tab:discrete_ablation} (c) reveals that $\stepratio = 5 \times 10^{-3}$ to be the optimal \rationame. This value was then used as the clamping value for our schedule.

Table \ref{tab:discrete_ablation}(d) shows that by combining these improvements, our discrete \modelname approach significantly reduces the performance gap between Shortcut models and the MeanFlow model.
\section{LLM usage}
\label{sec:llm-usage}

As requested by the ICLR 2026 policy\footnote{\url{https://iclr.cc/Conferences/2026/AuthorGuide}}, we disclose the usage of Large Language Models in this section.
LLMs were primarily used in two capacities:
\begin{itemize}
    \item Coding assistance for experiments. LLMs provided code auto-completion functionality to ease the process of implementing and analyzing the experiments.
    \item Writing assistance for paper writing. We used LLMs to assist with grammar and phrasing validation while working on the submission.
\end{itemize}

\section{Random vs balanced classes for FID computation}
\label{app:random_balance_class}

We treat EDM series~\citep{EDM, EDM2} as the standard in FID~\citep{FID} evaluations, which use a randomly sampled class label (from 0 to 999) for each sample in constructing 50,000 synthetic examples with the model.
We found a curious way to decrease the FID values by up to 10\% by using ``balanced'' class sampling: instead of using 50,000 independently sampled random classes, one can generate 50 samples for each of 1000 classes.
This greatly improves FID results, but not FDD (\ie, \Frechet Distance in the DINOv2~\citep{DINOv2} feature space) or FCD~\citep{FIDImageNetClasses} (\ie, \Frechet Distance in the CLIP-L-based~\citep{CLIP} feature space).

Since it is not a standard practice in the community, we only report it separately from the random class sampling results and with the appropriate notice.
But we emphasize that it might be a more reasonable way to evaluate FID since it reduces the variance (we are less likely to sample an unlucky set of classes).
We provide the results for it in \Cref{tab:balanced-fid}.

\begin{table*}[h]
\centering
\tablestyle{2pt}{1.}
\small
\begin{tabular}{lccc cccccc}
\toprule
\multirow{2}{*}{Method} & \multirow{2}{*}{Class sampling} & \multirow{2}{*}{Params} & \multirow{2}{*}{Epochs} & \multicolumn{3}{c}{NFE 1} & \multicolumn{3}{c}{NFE 2} \\
\cmidrule(lr){5-7} \cmidrule(lr){8-10}
& & & & \FID & \FDD & \FCD & \FID & \FDD & \FCD \\
\midrule
MeanFlow-XL/2$^{*}$ & Random $U[1..1000]$ & 676M & 240 & 3.47 & 185.8 & 3.39 & 2.46 & 108.7 & 2.40 \\
\modelname-XL/2~\ours & Random $U[1..1000]$ & 676M & 240 & 2.95 & 164.6 & 3.14 & 2.32 & 105.7 & 2.42 \\
\modelnamexlp~\ours & Random $U[1..1000]$ & 676M & 240+60 & \textbf{2.58} & \textbf{148.4} & \textbf{3.07} & $\textbf{2.15}$ & \textbf{96.8} & \textbf{2.31} \\
\midrule
MeanFlow-XL/2$^{*}$ & Balanced & 676M & 240 & 3.33 & 182.8 & 3.34 & 2.26 & 106.1 & 2.36 \\
\modelname-XL/2~\ours & Balanced & 676M & 240 & 2.81 & 162.4 & 3.10 & 2.16 & 103.2 & 2.37 \\
\modelnamexlp~\ours & Balanced&  676M & 240+60 & \textbf{2.44} & \textbf{147.2} & \textbf{3.04} & \textbf{1.95} & \textbf{94.6} & \textbf{2.30} \\
\bottomrule
\end{tabular}
\caption{\textbf{Balanced vs random class sampling for FID, FDD and FCD}.
}
\label{tab:balanced-fid}
\end{table*}

It is curious to observe that while it greatly improves FID results, FDD and FCD are barely affected.
We believe that this constitutes one more reason for the community to switch from FID to more robust metrics which correlate better with human perception, like FDD and FCD.

\clearpage
\section{Additional exploration of the MeanFlow loss}
\label{sec:additional-exploration}

\begin{figure}[h!]
\centering

\begin{subfigure}{0.45\textwidth}
\includegraphics[width=\linewidth]{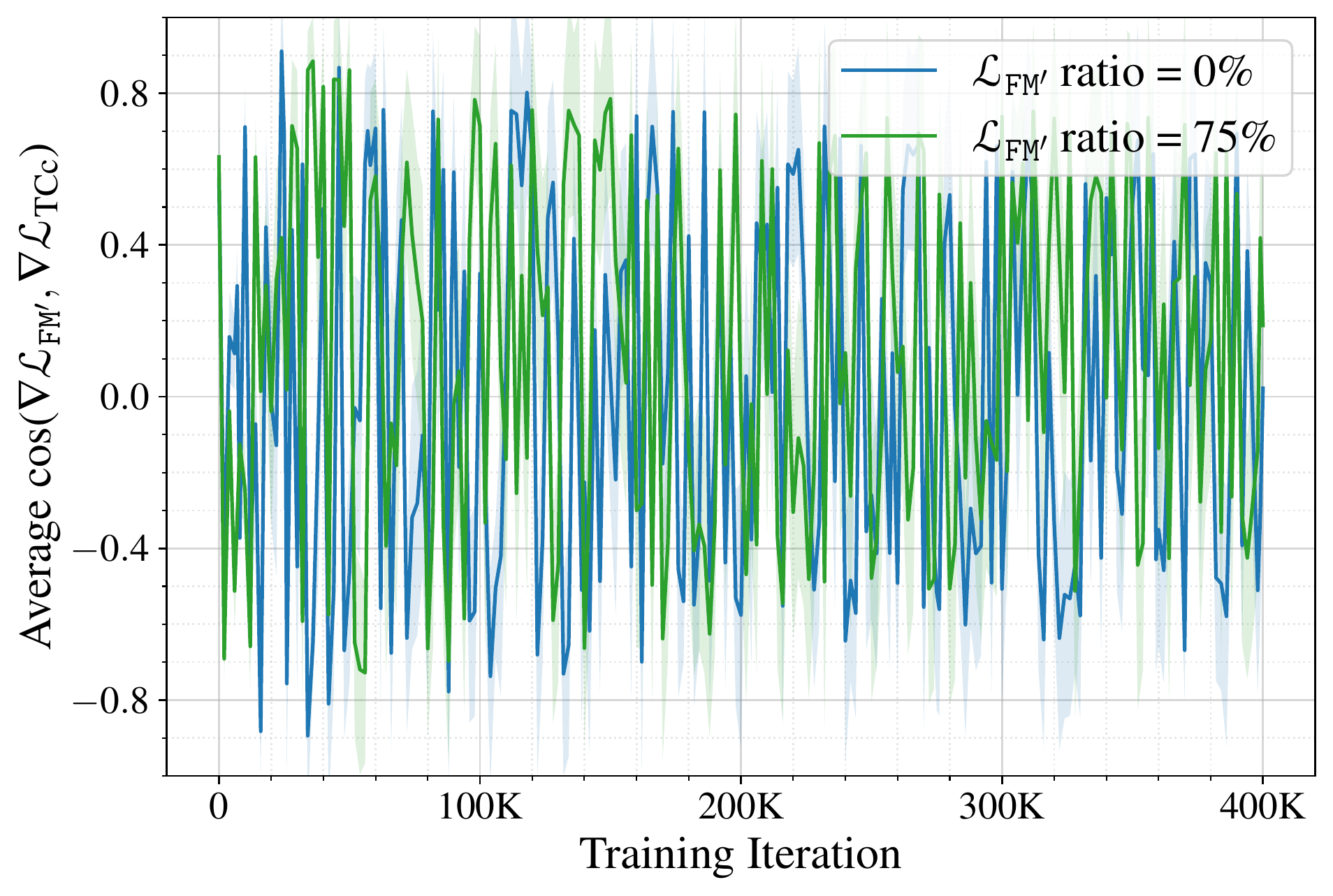}
\caption{$\cos\left(\nabla\LbFM, \nabla\LCFc\right)$}
\end{subfigure}\hfill
\begin{subfigure}{0.45\textwidth}
\includegraphics[width=\linewidth]{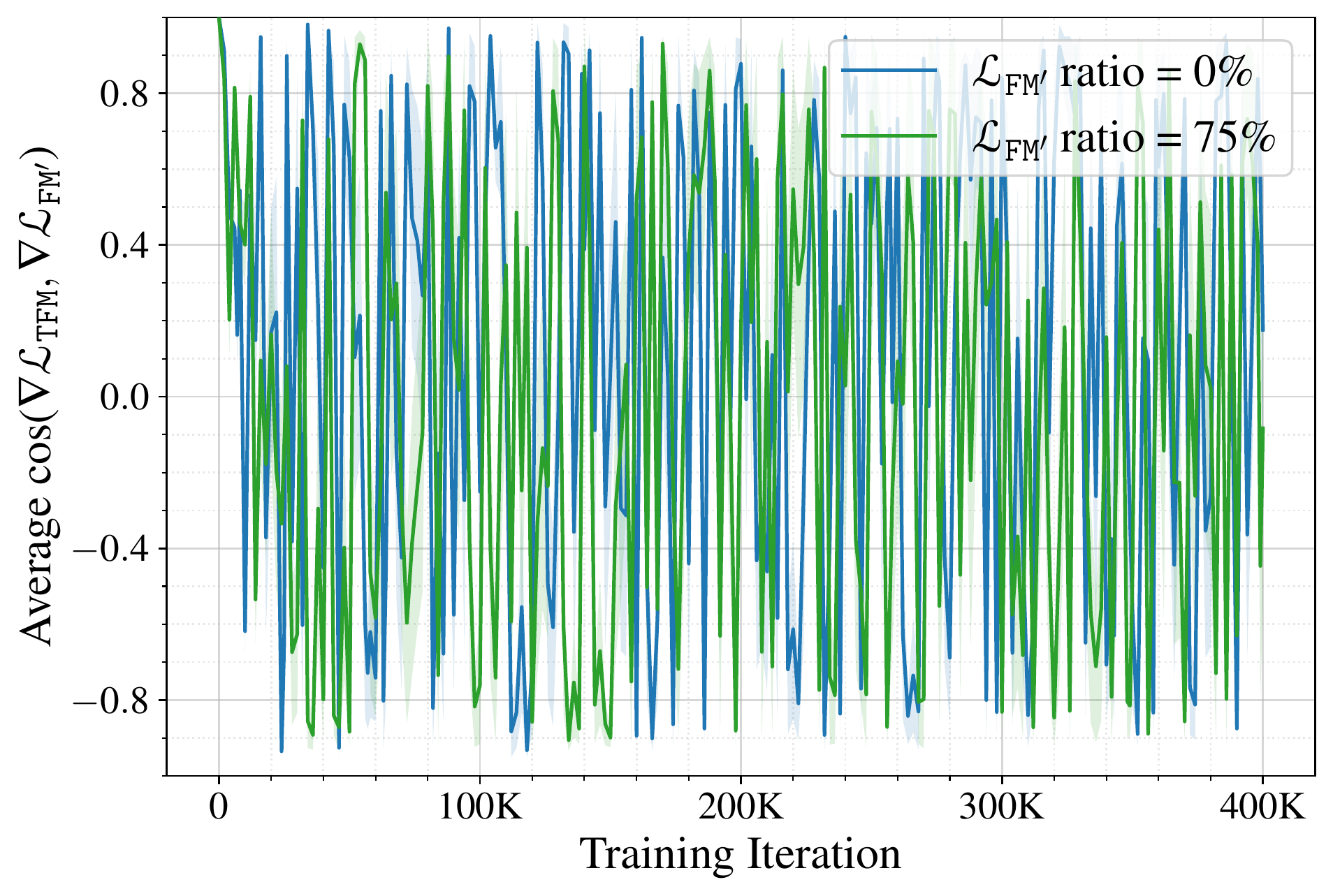}
\caption{$\cos\left(\nabla\LuFM, \nabla\LbFM\right)$}
\end{subfigure}

\begin{subfigure}{0.45\textwidth}
\includegraphics[width=\linewidth]{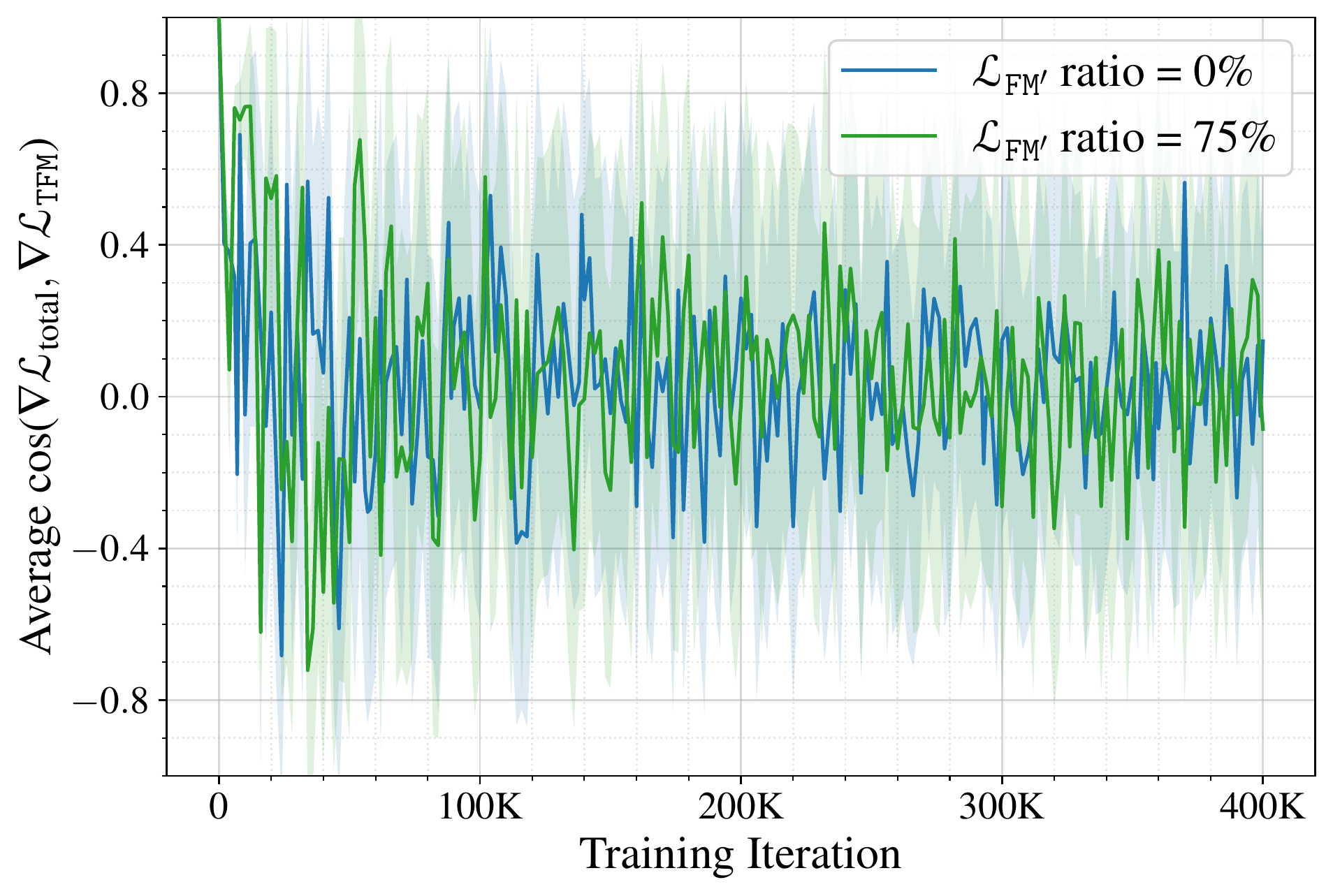}
\caption{$\cos\left(\nabla\LuFM, \nabla\LMFc\right)$}
\end{subfigure}\hfill
\begin{subfigure}{0.45\textwidth}
\includegraphics[width=\linewidth]{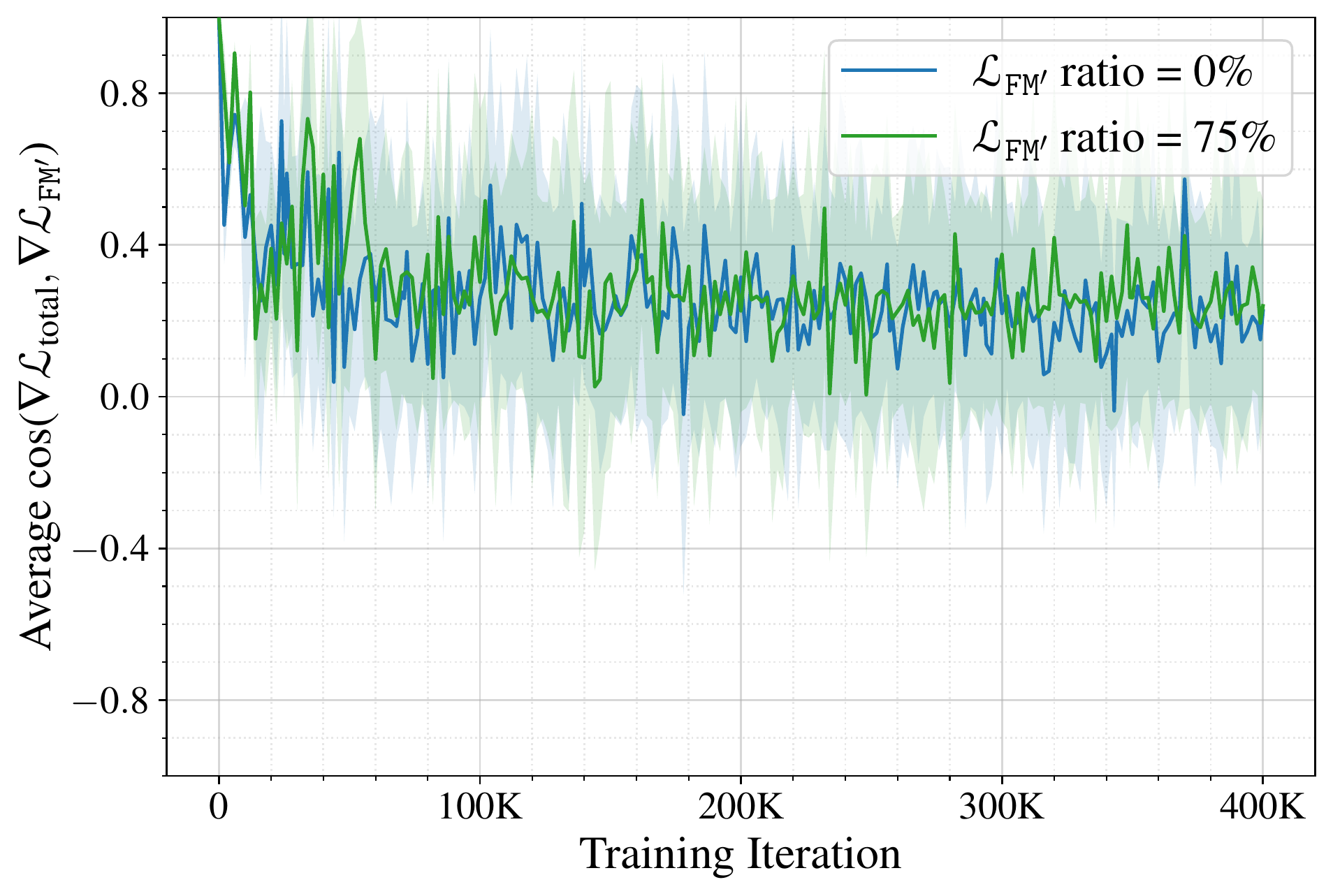}
\caption{$\cos\left(\nabla\LbFM, \nabla\LMFc\right)$}
\end{subfigure}

\begin{subfigure}{0.45\textwidth}
\includegraphics[width=\linewidth]{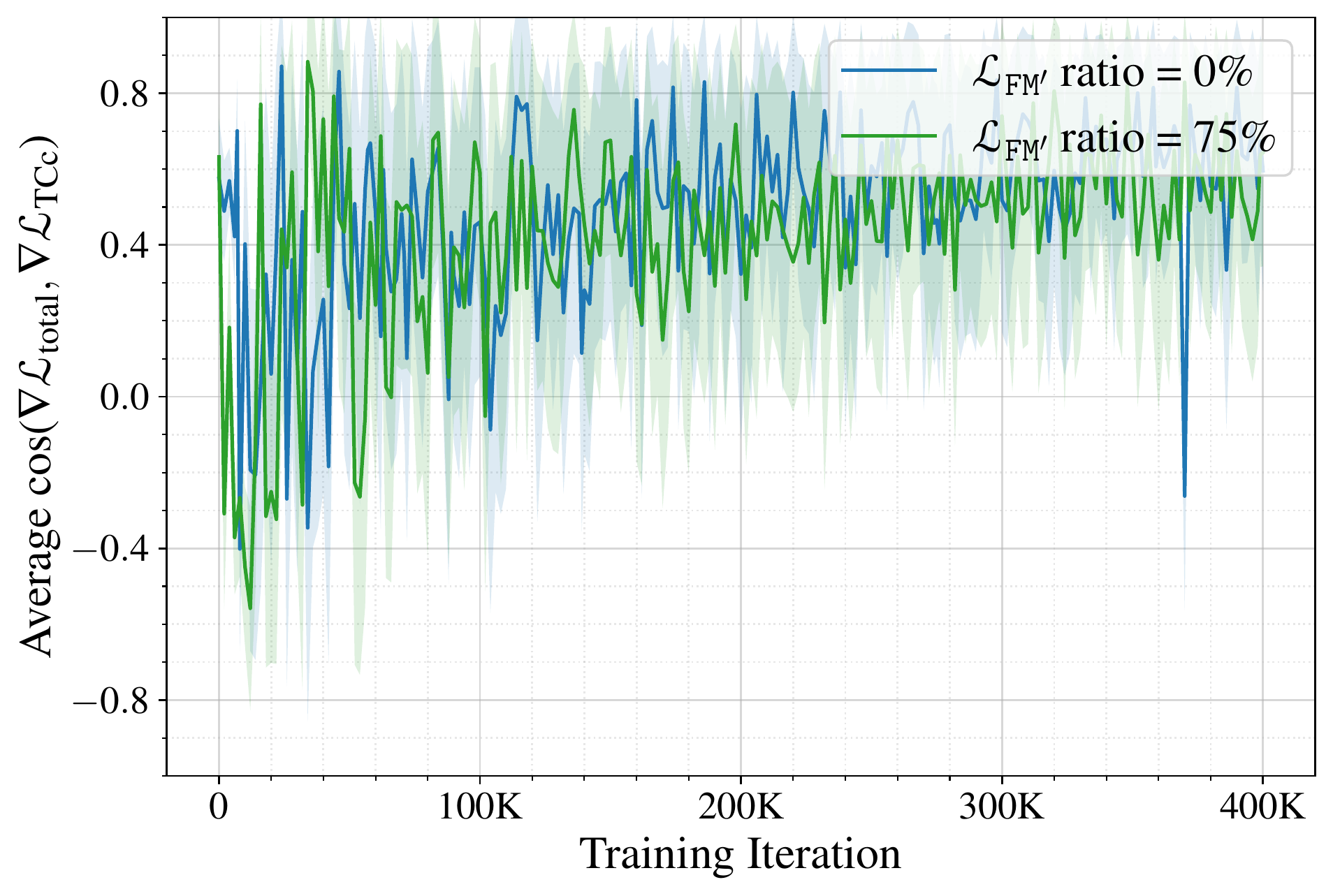}
\caption{$\cos\left(\nabla\LCFc, \nabla\LMFc\right)$}
\end{subfigure}\hfill
\begin{subfigure}{0.45\textwidth}
\includegraphics[width=\linewidth]{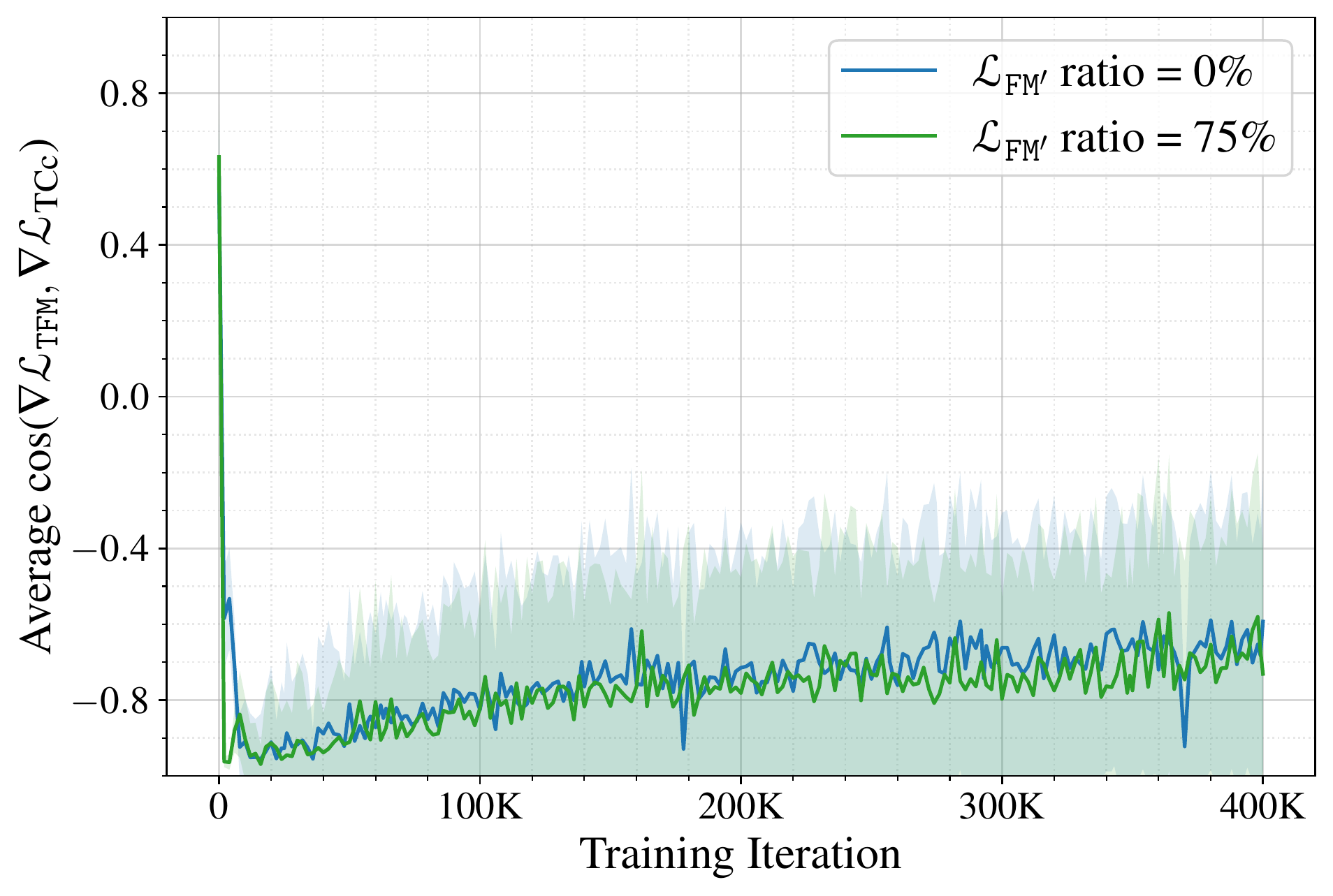}
\caption{$\cos\left(\nabla\LCFc, \nabla\LuFM\right)$}
\end{subfigure}
\caption{Average cosine similarities between the gradients of different losses ($\LuFM, \LbFM, \LCTc, \LMFc$) for DiT-B/2 MeanFlow model trained with 0\% and 75\% of flow matching.}
\label{fig:all-cos}
\end{figure}

\clearpage
\section{Additional visualizations}\label{app:additional_vis}

\newcommand{\suppsampleswidth}{0.9\linewidth}
\renewcommand{\arraystretch}{0.5}

\newcommand{\buildsamplesgridfigure}[3]{%
\vspace{-1em}
\begin{figure}[h]
\centering
\begin{tabular}{@{}m{1em}@{}m{\dimexpr\linewidth\relax}@{}}
\rotatebox{90}{MeanFlow-\ditxl} &
\includegraphics[width=\suppsampleswidth]{figure/jpg/ldm-2375-meanflow-ditxl2-class#1-ns#2-ts1.0.jpg} \\
\rotatebox{90}{$\alpha$-Flow-\ditxl} &
\includegraphics[width=\suppsampleswidth]{figure/jpg/ldm-2481-alphaflow-ditxl2-class#1-ns#2-ts1.0.jpg} \\
\rotatebox{90}{$\alpha$-Flow-\ditxl+} &
\includegraphics[width=\suppsampleswidth]{figure/jpg/ldm-2523-alphaflow-ditxl2-pp-class#1-ns#2-ts1.0.jpg} \\
\end{tabular}
\vspace{-0.7em}
\caption{Uncurated samples (seeds 1-16) for Class #1 (\protect#3) for NFE=#2.}
\label{fig:extra-samples:class-#1:nfe-#2}
\end{figure}
}

\buildsamplesgridfigure{15}{1}{robin}
\buildsamplesgridfigure{15}{2}{robin}
\buildsamplesgridfigure{29}{1}{axolotl}
\buildsamplesgridfigure{29}{2}{axolotl}
\buildsamplesgridfigure{33}{1}{loggerhead}
\buildsamplesgridfigure{33}{2}{loggerhead}
\buildsamplesgridfigure{88}{1}{macaw}
\buildsamplesgridfigure{88}{2}{macaw}
\buildsamplesgridfigure{89}{1}{cockatoo}
\buildsamplesgridfigure{89}{2}{cockatoo}
\buildsamplesgridfigure{127}{1}{white stork}
\buildsamplesgridfigure{127}{2}{white stork}
\buildsamplesgridfigure{279}{1}{arctic fox}
\buildsamplesgridfigure{279}{2}{arctic fox}
\buildsamplesgridfigure{980}{1}{volcano}
\buildsamplesgridfigure{980}{2}{volcano}
\buildsamplesgridfigure{975}{1}{lakeside}
\buildsamplesgridfigure{975}{2}{lakeside}

\renewcommand{\arraystretch}{1.0}

\end{document}